\documentclass[10pt,twocolumn,letterpaper]{article}

\usepackage{wacv}      %

\usepackage{graphicx}
\usepackage{amssymb}
\usepackage{booktabs}
\usepackage{algorithm2e}
\usepackage{graphicx}
\graphicspath{{figures/}}
\DeclareGraphicsExtensions{.pdf,.png,.jpg}
\usepackage{mathtools}%
\usepackage{tikz}
\usetikzlibrary{arrows, automata, bayesnet}
\usepackage{ulem}
\usepackage{subcaption}
\usepackage{comment}
\usepackage{dblfloatfix}

\newcommand{\independent}{\mbox{${}\perp\mkern-11mu\perp{}$}}

\newcommand{\argmax}{\mathop{\mathrm{argmax}}}

\usepackage[pagebackref,breaklinks,colorlinks]{hyperref}

\usepackage[capitalize]{cleveref}
\crefname{section}{Sec.}{Secs.}
\Crefname{section}{Section}{Sections}
\Crefname{table}{Table}{Tables}
\crefname{table}{Tab.}{Tabs.}

\def\wacvPaperID{2178}
\def\confName{WACV}
\def\confYear{2024}

\begin{document}

\newcommand{\MNAME}{CoPA}
\title{Robust Learning via Conditional Prevalence Adjustment} %
\author{Minh Nguyen$^{1}$\qquad Alan Q. Wang$^{1}$\qquad Heejong Kim$^{2}$ \qquad Mert R. Sabuncu$^{1,2}$\\
$^{1}$ Cornell University \\
$^{2}$ Department of Radiology, Weill Cornell Medical
}
\maketitle

\begin{abstract}
Healthcare data often come from multiple sites in which the correlations between confounding variables can vary widely.
If deep learning models exploit these unstable correlations, they might fail catastrophically in unseen sites.
Although many methods have been proposed to tackle unstable correlations, each has its limitations.
For example, adversarial training forces models to completely ignore unstable correlations, but doing so may lead to poor predictive performance.
Other methods (e.g.~Invariant risk minimization~\cite{arjovsky2019invariant}) try to learn domain-invariant representations that rely only on stable associations by assuming a causal data-generating process (input $X$ causes class label $Y$).
Thus, they may be ineffective for anti-causal tasks ($Y$ causes $X$), which are common in computer vision.
We propose a method called \MNAME~(Conditional Prevalence-Adjustment) for anti-causal tasks.
\MNAME~assumes that (1) generation mechanism is stable, i.e.~label $Y$ and confounding variable(s) $Z$ generate $X$, and (2) the unstable conditional prevalence in each site $E$ fully accounts for the unstable correlations between $X$ and $Y$.
Our crucial observation is that confounding variables are routinely recorded in healthcare settings and the prevalence can be readily estimated, for example, from a set of $(Y,Z)$ samples (no need for corresponding samples of $X$).
\MNAME~can work even if there is a single training site, a scenario which is often overlooked by existing methods.
Our experiments on synthetic and real data show \MNAME~beating competitive baselines.
\end{abstract}

\section{Introduction}
Out-of-domain (OOD) generalization is essential in many fields like~healthcare, in which data come from multiple sites.
Between sites, the data are not identically distributed, and correlations between (confounding) variables can vary widely (i.e., are unstable).
For example, different hospitals may use different imaging devices, making the scans look different.
Furthermore, imaging techniques may be spuriously correlated with diagnosis at some hospitals but not others.
ML models trained to diagnose using images might exploit unstable correlations~\cite{albadawy2018deep,pooch2020can,degrave2021ai} to increase training predictive accuracy and could perform poorly at new sites.

\begin{figure*}[t]
\centering
\includegraphics[width=\linewidth]{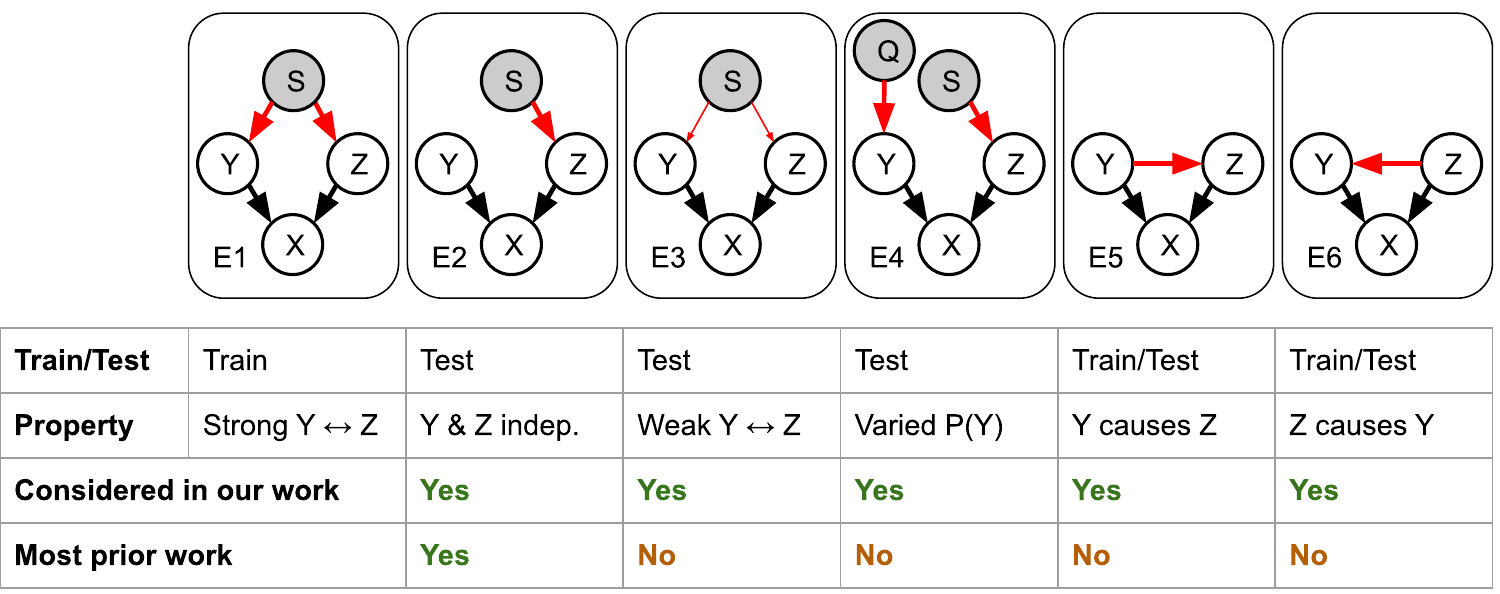}
\caption{Data generation at different sites with the same stable generative distribution $P(X|Y,Z)$.
Red edges are unstable (i.e.~generative mechanisms vary with sites) while black edges are stable.
$X$, $Y$, $Z$ are input, target, and confounding variables respectively.
Most methods assume strong spurious correlation in training data (E1) which vanishes in test data (E2).
However, that is not the only possibility.
Others include: weak spurious correlation (E3), label-shift (E4), or causal correlation (E5 and E6).
Gray nodes are hidden/unobserved.
}\label{fig:scenarios}
\end{figure*}

Understanding the data-generating process and how it changes between sites could help account for unstable correlations.
In this work, we restrict our attention to the case where the label $Y$ (e.g., an object in a scene) and confounding variable(s) $Z$ (e.g., camera type) are causes of $X$ (e.g., the image).
$Y$ and $Z$ may be (spuriously) correlated.
Figure~\ref{fig:scenarios} shows 6 causal graphs ($E1$ through $E6$) representing 6 data-generating processes under this case.
Unstable correlations are indicated with red edges in the graphs.
We also assume that the mechanism that generates $X$ from its causal parents (i.e.~$P(X|Y,Z)$) is stable, while the remaining mechanisms may vary between sites.
Consequently, the correlations between $X$ and its parents are stable and are denoted using black edges.

Some prior methods, like~domain-adversarial training~\cite{ganin2016domain}, aim to ensure that the model does not exploit spurious correlations between $Y$ and $Z$.
Such methods implicitly assume that the unstable correlations between $Y$ and $Z$ (through the backdoor path) can vanish in test data, as shown in $E2$.
When the test data distribution deviates from $E2$, however, these methods can be sub-optimal.
For example, the unstable correlations between $Y$ and $Z$ may simply change in degree (e.g., weaken as in $E3$), so exploiting these may still be useful for predictions~\footnote{Consider a scenario where patients are triaged based on risk factors correlated with diagnosis $Y$ and imaging parameters $Z$. $E1$ and $E3$ can correspond to different triaging systems.}.
Additionally, data generation can change due to a changing prior on $Y$, i.e., label-shift ($E4$).
Although some methods have been proposed to address label-shift~\cite{lipton2018detecting}, it remains an under-studied problem~\cite{schrouff2022maintaining}.
There is a lack of methods that account for both spurious correlations and label-shift~\cite{schrouff2022maintaining} even though they often co-occur in reality.
Furthermore, these methods are also not applicable when the link between $Y$ are $Z$ are causal ($Y$ causing $Z$, as in $E5$ or vice-versa, as in $E6$).
Other methods, e.g.~Invariant risk minimization (IRM)~\cite{arjovsky2019invariant}, leverage data from multiple training sites to extract a domain-invariant representation, which is assumed to be transportable to any site.
IRM learns to predict $Y$ using some representation $\Phi$ of $X$ that is a function of the causal parents (PA) of $Y$.
More precisely, IRM learns functions $f$ and $\Phi$ such that $\widehat{Y}{:=}\argmax_{Y} P(Y|\mathit{PA}(Y)){:=}f(g(\mathit{PA}(Y))){:=}f(\Phi(X))$.
By assuming that $P(Y|\mathit{PA}(Y))$ is stable (domain-invariant), IRM is also stable and it will perform well in all sites.
However, IRM is not formulated for anti-causal learning problems ($Y$ is an ancestor of input $X$) because $\Phi(X)$ cannot be some function of $\mathit{PA}(Y)$.
Consequently, using IRM in anti-causal problems (very common in computer vision~\cite{scholkopf2012causal}) can result in bad OOD performance~\cite{ahuja2021invariance}, especially in the presence of label-shift~\cite{wang2022cisa}.
Besides, IRM and its variants rely on training data from multiple sites, which may be possible to obtain.

We propose an approach for anti-causal learning named Robust learning via Conditional Prevalence-Adjustment, or \MNAME~for short.
\MNAME~learns a stable predictor~\cite{subbaswamy2022unifying} of $Y$ that leverages the stable edges and an estimate of the conditional prevalence $P(Y|Z,E)$ in each site $E$.
By adjusting for the effect of unstable correlations through the conditional prevalence estimate, \MNAME~can learn to generalize to OOD samples.
Crucially, the conditional prevalence estimate at each site, including test sites, can be readily obtained from a set of $(Y,Z)$ samples without any need for labeled samples of $X$.
This estimation is helped by the fact that confounding variables $Z$ are routinely recorded in healthcare ($Z$ are visible/observed).
\MNAME~has several advantages over baselines.
\begin{itemize}
    \item Since the conditional prevalence estimate absorbs the effect of label-shift, \MNAME~is less susceptible to this change which is quite common in healthcare data (e.g.~disease prevalence can vary between hospitals).
    \item \MNAME~can deal with not only spurious correlations (Figure~\ref{fig:scenarios}, $E1$ to $E4$) but also changing causal correlations (Figure~\ref{fig:scenarios}, $E5$ and $E6$) because the prevalence-adjustment procedure of \MNAME~does not assume any specific causal ordering between $Y$ and $Z$.
    \item \MNAME~can work even if there is a single training site, a scenario sometime overlooked by existing methods.
\end{itemize}
Our experiments on synthetic and real data show \MNAME~outperforming competitive baselines and demonstrates good OOD generalization.

\section{Related Work}\label{sec:related}
In OOD settings where data is assumed to be available from multiple sites and the sites are known, there are several frameworks with different assumptions ~\cite{gulrajani2021search}.
Domain adaptation assumes access to test sites' unlabeled data~\cite{pan2010survey}.
Transfer learning assumes access to some labeled data from test sites~\cite{wilson2020survey}.
Domain generalization assumes no information of test sites is available~\cite{peters2016causal,arjovsky2019invariant}.
Our setup assumes access to some statistics of class labels from the test sites, thereby most resembling domain generalization.

Domain-invariant representation learning~\cite{muandet2013domain,li2018domain,zhao2019learning,tanwani2021dirl} aims to learn an invariant representation across multiple domains to achieve better OOD generalization.
One could apply domain-invariant representation learning via adversarial learning~\cite{ganin2016domain,li2018deep} for domain generalization.
However, these methods may fail in the presence of label-shift~\cite{arjovsky2019invariant,zhao2019learning,tachet2020domain}.

IRM~\cite{arjovsky2019invariant} is another approach to domain generalization which learns invariant causal predictors~\cite{peters2016causal} using data from multiple sites.
However, IRM may fail when (1) there are too few training sites~\cite{rosenfeld2021risks}, (2) the number of samples per site is too low~\cite{kamath2021does}, or (3) when test sites are very different from training sites~\cite{rosenfeld2021risks}.
Follow-up work such as Risk Extrapolation (REx)~\cite{krueger2021out} have been proposed to tackle more extreme shifts between training and test sites.
Yet, the requirement for multiple training sites still remains.
Other notable methods for domain generalization include CORAL~\cite{sun2016deep} and DRO~\cite{sagawa2019distributionally}.
Unfortunately, few can consistently beat ERM in real-world settings~\cite{gulrajani2021search}.
More recent methods such as IWDANN~\cite{tachet2020domain} and LAMDA~\cite{le2021lamda} try to tackle both domain adaptation and the label-shift problem.
However, they were formulated for only 2 sites (1 source and 1 target).
Construction of realistic benchmarks such as the WILDS benchmark~\cite{koh2021wilds} has been beneficial for domain generalization research.
However, these benchmarks currently lack information about potential confounders and they do not consider label-shift.

\section{Proposed Method}\label{sec:method}
\MNAME~assumes (1) a stable mechanism for generating $X$ from label $Y$ and confounders $Z$; (2) the availability of the conditional prevalence $P(Y|Z,E)$ at each site $E$; and (3) the observability of confounders $Z$ at training and test sites.
Since confounders are routinely collected in healthcare, the second and third assumptions usually hold.
Nevertheless, we explore how to relax these assumptions in Section~\ref{sec:ablate}.

\subsection{Conditional Prevalence Adjustment Across Sites}\label{ssec:method_prevalence}
Since $P(X|Y,Z)$ is assumed to be stable (i.e.~invariant across sites), $X\independent E|Y,Z$.
For brevity, we denote $P(\cdot|\cdot,E{=}e)$ as $P(\cdot|\cdot,e)$.
For any two sites $e_i$ and $e_j$:
\begin{align}
    P(X|Y,Z,e_i) &= P(X|Y,Z) = P(X|Y,Z,e_j) \label{eq:stability} \\
    &= P(Y|X,Z,e_i)\frac{P(X|Z,e_i)}{P(Y|Z,e_i)}, \label{eq:bayes_exp}
    \end{align}
where \eqref{eq:bayes_exp} follows from Bayes' rule.
From~\eqref{eq:stability} and~\eqref{eq:bayes_exp}:
\begin{align}
    &P(Y|X,Z,e_j) = \frac{P(Y|Z,e_j)}{P(Y|Z,e_i)} \frac{P(X|Z,e_i)}{P(X|Z,e_j)} P(Y|X,Z,e_i) \label{eq:manip}
\end{align}
Using~\eqref{eq:manip}, the maximum-likelihood estimator of $Y$ given input $X$ and $Z$ at site $e_j$ can be expressed as
\begin{align}
\widehat{Y}_{e_j} &= \argmax_{Y} P(Y|X,Z,e_j) \nonumber\\
    &= \argmax_{Y} P(Y|Z,e_j)\frac{P(Y|X,Z,e_i)}{P(Y|Z,e_i)}. \label{eq:prevalence}
\end{align}
Let $R(X,Z)$ be the ratio $P(Y|X,Z,E)/P(Y|Z,E)$.
Equation~\eqref{eq:prevalence} implies that $R(X,Z)$ is invariant across sites and all the site-specific instability can be absorbed by the conditional prevalence $P(Y|Z,e_j)$.

This suggests a new domain adaptation strategy.
Let $f_\theta(X,Z)$ denote an estimator, with parameters $\theta$, which models the ratio $R(X,Z)$.
One can adapt the predictor to the new site by adjusting for the new site prevalence $P(Y|Z,e_j)$.
Specifically, if the predictor at site $e_i$ is:
\begin{align}
    & P(Y|X,Z,e_i) = P(Y|Z,e_i) f_\theta(X,Z)
\end{align}
then $P(Y|Z,e_j) f_\theta(X,Z)$ can be used to predict for samples at an unseen site $e_j$.

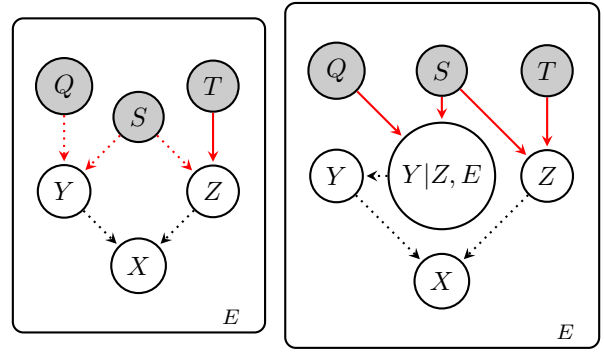
\begin{figure}[t]
\centering
\begin{minipage}{.40\columnwidth}
\centering
\begin{tikzpicture}[> = stealth, shorten > = 1pt, auto, node distance = 1.4cm, thick]
    \tikzstyle{every state}=[draw=black, thick, fill=white, minimum size=4mm]
    \node[state, fill=black!20] (S) {$S$};
    \node[state] (Y) [below left of=S]{$Y$};
    \node[state] (Z) [below right of=S] {$Z$};
    \node[state] (X) [below right of=Y] {$X$};
    \node[state, fill=black!20] (Q) [above of=Y] {$Q$};
    \node[state, fill=black!20] (T) [above of=Z] {$T$};
    \plate [inner sep=.3cm,xshift=.02cm,yshift=.2cm] {plate1} {(X) (Y) (Z) (Q) (T)} {$E$}; %
    \path[->, dotted, red, left] (Q) edge node {} (Y);
    \path[->, dotted, red, above] (S) edge node {} (Y);
    \path[->, dotted, red, below] (S) edge node {} (Z);
    \path[->, red] (T) edge node {} (Z);
    \path[->, dotted] (Y) edge node {} (X);
    \path[->, dotted] (Z) edge node {} (X);
\end{tikzpicture}
\end{minipage}%
\hspace{.2cm}
\begin{minipage}{.45\columnwidth}
\centering
\begin{tikzpicture}[> = stealth, shorten > = 1pt, auto, node distance = 1.4cm, thick]
    \tikzstyle{every state}=[draw=black, thick, fill=white, minimum size=4mm]
    \node[state, fill=black!20] (S) {$S$};
    \node[state, fill=black!20, left of=S] (Q)  {$Q$};
    \node[state, fill=black!20, right of=S] (T)  {$T$};
    \node[state] (YZE) [below of=S]{$Y|Z,E$};
    \node[state] (Y) [left of=YZE]{$Y$};
    \node[state] (Z) [right of=YZE] {$Z$};
    \node[state] (X) [below of=YZE] {$X$};
    \plate [inner sep=.3cm,xshift=.02cm,yshift=.2cm] {plate1} {(X) (Y) (Z) (Q) (T)} {$E$}; %
    \path[->, red, left] (Q) edge node {} (YZE);
    \path[->, red, left] (S) edge node {} (YZE);
    \path[->, red] (T) edge node {} (Z);
    \path[->, red, right] (S) edge node {} (Z);
    \path[->, dotted] (YZE) edge node[left] {} (Y);
    \path[->, dotted] (Y) edge node[below] {} (X);
    \path[->, dotted] (Z) edge node {} (X);
\end{tikzpicture}
\end{minipage}%
\caption{A predictor of $Y$ using input $X$ and $Z$ leverages unstable edges (left).
A predictor of $Y$ using $X$, $Z$, and $P(Y|Z,E)$ as input only uses stable edges, hence it is a stable predictor (right).
Dotted edges: statistical relations used in predictors.
Black edges: stable, red edges: unstable.
White nodes: visible, gray nodes: hidden.
}\label{fig:prevalence_intuition}
\end{figure}

\subsubsection{Additional Intuition}
Figure~\ref{fig:prevalence_intuition} provides additional intuition for \MNAME.
Given the graph (Figure~\ref{fig:prevalence_intuition}, left panel), the statistical relations (links in causal graph) used by $P(Y|X,Z,E)$ and $P(Y|Z,E)$ are:
\begin{itemize}
    \item $P(Y|X,Z,E)$: $Q{\rightarrow}Y$, $Y{\rightarrow}X$, $Z{\rightarrow}X$, $Y{\leftarrow}S{\rightarrow}Z$
    \item $P(Y|Z,E)$: $Q{\rightarrow}Y$ and $Y{\leftarrow}S{\rightarrow}Z$
\end{itemize}
Specifically, $Y{\rightarrow}X$ and $Z{\rightarrow}X$ are used to infer $Y$ from $X$; $Q{\rightarrow}Y$ is used to infer $Y$ from $E$; and the back-door path $Y{\leftarrow}S{\rightarrow}Z$ is used to infer $Y$ from $Z$.
From Equation~\ref{eq:prevalence}, $P(Y|X,Z,e_j)$ is the product of $P(Y|Z,e_j)$ and the ratio $R(X,Z)$.
Furthermore, since $P(Y|X,Z,e_j)$ uses 4 links and $P(Y|Z,e_j)$ already accounts for 2 links, $R(X,Z)$ only needs to account for the remaining 2 links, namely $Y{\rightarrow}X$ and $Z{\rightarrow}X$.
Consequently, $R(X,Z)$ is invariant across sites because $Y{\rightarrow}X$ and $Z{\rightarrow}X$ are stable (due to the stable generation assumption).

Since the ratio is invariant, the instability of $P(Y|X,Z,e_i)$ is captured in the term $P(Y|Z,e_i)$.
Thus, when $P(Y|Z,e_i)$ is known, this effectively shields the prediction of $Y$ from site instability (shown in Figure~\ref{fig:prevalence_intuition}, right panel).
Hence, one can construct a predictor of $Y$ from $X$, $Z$, and $P(Y|X,Z,e_i)$ that is domain-invariant.
As the instability captured in $P(Y|Z,e_i)$ includes the label-shift effect on $Y$ due to $Q$ and $S$, prevalence-adjustment makes \MNAME~robust to label-shift.
Furthermore, since the above argument for prevalence-adjustment can be adapted to cases where the link between $Y$ and $Z$ is causal ($Y$ causes $Z$ or $Z$ causes $Y$) instead of spurious, without loss of generality, \MNAME~can be applied to other sites (e.g.~$E5$ and $E6$ in Figure~\ref{fig:scenarios}) even when the exact causal relation between $Y$ and $Z$ is not known.
\begin{algorithm}[t]
\caption{\MNAME. $L_{\mathit{Ent}}$: cross-entropy loss}\label{alg:implicit}
\hrulefill\\
\KwIn{\\
    $D_{\mathit{train}}$:   $\{ x^e_k, y^e_k, z^e_k\}, \widehat{P}(Y|Z,e) , \forall e\in\{e_1,..,e_t\}$\\
    $D_{\mathit{test}}$:   $\{ x^e_k, z^e_k\}, \widehat{P}(Y|Z,e) , \forall e\in\{e_{t{+}1},..,e_N\}$
    }
\KwOut{$\{\widehat{y}^e_k\}, \forall e\in\{e_{t{+}1},..,e_N\}$}

1. Initialize neural network $f_\theta(X,Z)$\;
2.
\While{not converged}{
    \ForAll{$x^e_k, y^e_k, z^e_k$ in $D_{\mathit{train}}$}{
        $\hat{y}^e_k = \widehat{P}(Y|z^e_k,e) \odot f_\theta(x^e_k, z^e_k)$\;
        $\mathit{L} = L_{\mathit{Ent}}(y^e_k , \hat{y}^e_k)$\;
        Back-propagate $\mathit{L}$ and update $f_\theta$
    }
}
3.
\ForAll{$x^e_k, z^e_k$ in $D_{\mathit{test}}$}{
    $ \widehat{y}^e_k = \widehat{P}(Y|z^e_k,e) \odot f_\theta(x^e_k, z^e_k)$
}
\hrulefill\\
\end{algorithm}

\subsection{The \MNAME~Algorithm}\label{ssec:algorithm}
In \MNAME, we implement a model $f_\theta(X,Z)$ that captures the invariant ratio $R(X,Z)$.
In each site $e_i$, the site-specific conditional distribution of $Y$ is obtained by multiplying the output of $f_\theta(X,Z)$ with the site-specific prevalence $P(Y|Z,e_i)$.
This output is then compared against the ground-truth to calculate the gradients for model training.
We use cross-entropy as the loss function.

\begin{figure}[t]
\centering
\includegraphics[width=\linewidth]{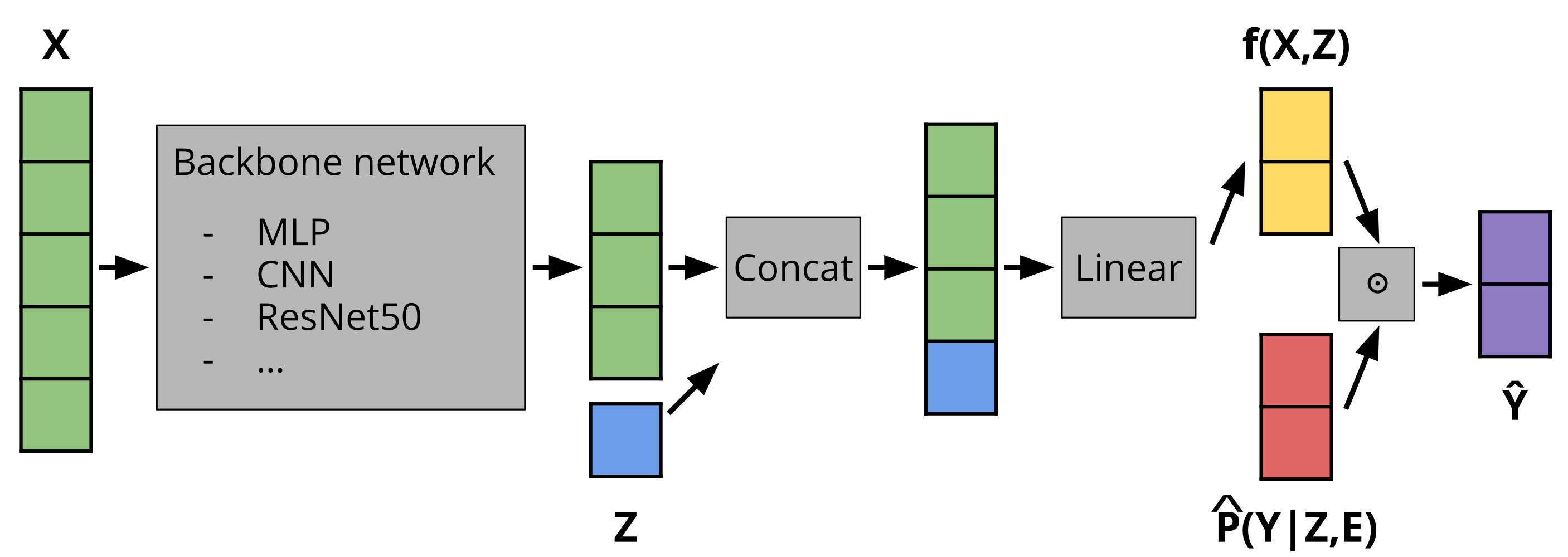}
\caption{Model architecture of \MNAME.}\label{fig:arch}
\end{figure}

Algorithm~\ref{alg:implicit} summarizes the steps in \MNAME.
Step 1 initializes the neural network, $f_\theta(X,Z)$, which is shown in Figure~\ref{fig:arch}.
Step 2 trains $f_\theta(X,Z)$ using gradient descent until convergence.
Model selection is performed according to validation criteria discussed in Section~\ref{ssec:validation}.
Step 3 uses the network to predict the labels of samples at new sites.

When $Z$ is a categorical variable, the smoothed empirical normalized counts can be used as the conditional prevalence estimates $\widehat{P}(Y|Z,E)$ (see Section~\ref{ssec:prevalence_estimate} for more details).
When $Z$ is a continuous variable or multi-dimensional, the empirical conditional prevalence estimate can be obtained by multiple training auxiliary models, one for each site $E$, to predict the probability of $Y$ given $Z$.

\subsection{Network Architecture}
Figure~\ref{fig:arch} shows the general architecture of the \MNAME~model which uses $X$, $Z$, and $\widehat{P}(Y|Z,E)$ to predict $Y$.
First, the representation of $X$ is computed using the backbone network.
This representation of $X$ is then combined with $Z$ via concatenation (late fusion) and the concatenated vector is fed through a linear layer.
The output of this linear layer is the domain-invariant ratio $f_\theta(X,Z) := R(X,Z)$.
Since this ratio is non-negative, the activation after the linear layer must be appropriately chosen.
In practice, we found that taking the softmax of the last layer worked well.
The output $f_\theta(X,Z)$ is then element-wise multiplied with the prevalence estimate, $\widehat{P}(Y|Z,E)$, to produce the conditional distribution $\widehat{P}(Y|X,Z,E)$.
The predicted label is the most likely class (argmax) of $\widehat{P}(Y|X,Z,E)$.

\section{Experiments}\label{sec:experiments}
We conducted experiments using both synthetic (Section~\ref{ssec:syn_exp}) and real data (Section~\ref{ssec:real_exp}).
Examples of the synthetic and real data are shown in Appendix~\ref{app:data_example}.
We experiment on the following scenarios to accurately reflect those that may arise in reality:
\begin{enumerate}
    \item \textit{Multiple vs single training site(s)}. First, while models trained on data from multiple sites may achieve better OOD performance, sometimes only data from a single site (e.g., hospital) might be available.
    Hence, it is important that methods can perform well in the single training site setup.
    \item \textit{Different causal relations between $Y$ and $Z$}.
    In some cases, the causal relations between the target $Y$ and the confounding variable $Z$ are not clearly understood.
    Thus, methods which can work regardless of the nature of the relationship between $Y$ and $Z$ are desirable.
\end{enumerate}

\subsection{Baselines}\label{ssec:baselines}
We compared \MNAME~against Empirical Risk Minimization~\footnote{ERM is the standard approach used in machine learning where one ignores the sites and minimizes the average loss over the training data.} (ERM) and four strong baselines for robust learning: IRM~\cite{arjovsky2019invariant}, DANN~\cite{ganin2016domain}, CORAL~\cite{sun2016deep}, and DRO~\cite{sagawa2019distributionally}, and IWDANN~\cite{tachet2020domain}.
IWDANN (Importance-Weighted DANN) was originally formulated for 2 sites but we extended IWDANN to the multi-site setup by following the authors' suggestion of having one set of importance weights for each pair of sites.
CORAL, DANN, and IWDANN have additional access to unlabeled data from validation and test sets.
For experiments with multiple training sites, we cycle through the sites between batches.
IRM is excluded in experiments with a single training site as it needs data from multiple sites.
For each method, results from 5 different runs using different random seeds were averaged.
Standard errors over these runs are indicated with error bars in the figures.
Given the unbalanced label distributions, F1-score instead of accuracy is used to evaluate performance.

\subsection{Estimating Empirical Prevalence}\label{ssec:prevalence_estimate}
When both $Y$ and $Z$ are categorical variables, the empirical prevalence $\widehat{P}(Y|Z,e_i)$ can be calculated directly by counting.
This is the case for synthetic data.
Our simulation created a separate set of $(Y,Z)$ labels in each site.
Let $L_i$ be the set of $(Y,Z)$ pairs used for prevalence estimation for site $e_i$.
The empirical prevalence $\widehat{P}(Y=y|Z=z,e_i)$ is simply the ratio
$\frac{\sum_{(Y,Z)\in L_i}\mathbb{I}[Y=y;Z=z]} {\sum_{(Y,Z)\in L_i}\mathbb{I}[Z=z]}$, where $\mathbb{I}$ is the indicator function.
For real data, there are multiple confounders $Z$ and some of them may be continuous.
Instead of counting, the empirical prevalence estimate can be obtained by training auxiliary models, one for each site $e_i$, to predict the probability of $Y$ given input $Z$.
Since the real datasets used in this paper do not include separate sets of $(Y, Z)$ samples, we have to use the same data for training/testing and prevalence estimation.
To avoid label leakage from prevalence estimation, the $(Y, Z)$ samples for a site is split into two halves and the fitted model using data from one half is used to estimate $\widehat{P}(Y|Z,e_i)$ for samples from the other half.

\subsection{Validation}\label{ssec:validation}
For all approaches, the best models during training are selected for evaluation on the OOD test data.
Model selection could try to (1) minimize in-domain validation error or (2) minimize the model's instability to distribution shifts~\cite{wald2021calibration}.
We measure the latter using validation error on data from an unseen site (termed \textit{external} validation).
We measure the former on held-out validation data from training sites (termed \textit{internal} validation).
The number of samples used to estimate \textit{internal} and \textit{external} validation error are kept equal.
The results presented in Section~\ref{sec:experiments} are based on \textit{external} validation.
Evaluation results using \textit{internal} validation are included in Appendix~\ref{app:complete_results}.

\begin{table}[t]
\setlength{\tabcolsep}{4pt}
\centering
\begin{tabular}{llll}
    Setup     &  Train  &  Val. & Test \\
    \midrule
    Multiple  &  (10k, 0.9), (10k, 0.7) &  (0.5k, 0.5) & (1k, 0.3) \\
    Single    &  (20k, 0.9)             &  (0.5k, 0.5) & (1k, 0.3)
\end{tabular}
\caption{Training, validation (\textit{external}), and test data in two different setups.
Each pair of numbers, $(N,\beta)$, represents a site with $N$ data samples generated using coefficient $\beta$.
}\label{tab:syn_data}
\end{table}
\subsection{Synthetic Data Experiments}\label{ssec:syn_exp}
\subsubsection{Data}\label{sssec:syn_data}
The $Y$ and $Z$ labels of the synthetic data were generated according to Equation~\ref{eq:sem_begin}-\ref{eq:sem_end}.
There are 3 different setups corresponding to 3 different causal relations between $Y$ and $Z$.
$\mathsf{Unif}(0,1)$ denotes a uniform random variable on $(0,1)$, and
$\mathsf{Norm}(\mu, \sigma^2)$ is a Gaussian with mean $\mu$ and variance $\sigma^2$.
The value of $\alpha$ is set at 0.3.
$\beta$ is a site-specific coefficient within the range $(0, 1)$.
Larger $\beta$ corresponds to a stronger correlation between $Y$ and $Z$.
As $\beta$ varies, the $Y$ label distribution also shifts.
$Y$ and $Z$ are binary variables.
\begin{align}
    \shortintertext{Common cause (Figure~\ref{fig:scenarios}, $E1/E2/E3/E4$)}
    S &\leftarrow \mathsf{Unif}(0,1) \label{eq:sem_begin}\\
    Y &\leftarrow \mathbb{I}\big[ \beta S + (1-\beta)\alpha > 0.5 \big] \\
    Z &\leftarrow \mathbb{I}\big[ \beta S + (1-\beta)\mathsf{Unif}(0, 1) > 0.5 \big] \\
    \shortintertext{$Y$ causes $Z$ (Figure~\ref{fig:scenarios}, $E5$)}
    Y &\leftarrow \mathbb{I}\big[ \beta\mathsf{Unif}(0, 1) + (1-\beta)\alpha > 0.5 \big] \\
    Z &\leftarrow \mathbb{I}\big[ \beta Y/2 + (1-\beta/2)\mathsf{Unif}(0, 1) > 0.5 \big] \\
    \shortintertext{$Z$ causes $Y$ (Figure~\ref{fig:scenarios}, $E6$)}
    Z &\leftarrow \mathbb{I}\big[ \mathsf{Unif}(0, 1) > 0.5 \big] \\
    Y &\leftarrow \mathbb{I}\big[ \beta Z/2 + \beta\mathsf{Unif}(0, 1)/2 + (1-\beta)\alpha > 0.5 \big] \label{eq:sem_end}
\end{align}
\\
We consider two types of synthetic $X$: 2-dim and CMNIST. \\
\textbf{2-dim:} The first type is low-dimensional where the input $X$ is a 2-dim vector generated from target $Y$ and an auxiliary variable $Z$ is correlated with $Y$ according to Equations~\ref{eq:2bit_begin}-\ref{eq:2bit_end}.
$W\in\mathbb{R}^{2\times 2}$ denotes a randomized mixing matrix that is the same (stable) across different sites.
\begin{align}
    C_1 &\leftarrow 0.1\mathbb{I}[Y=1] -0.1\mathbb{I}[Y=0] + \mathsf{Norm}(0, 0.1^2) \label{eq:2bit_begin}\\
    C_2 &\leftarrow 1.0\mathbb{I}[Z=1] -1.0\mathbb{I}[Z=0] + \mathsf{Norm}(0, 0.1^2) \\
    X &\leftarrow W \times [C_1, C_2]\label{eq:2bit_end}
\end{align}
\\
\textbf{CMNIST:}
The second type is higher-dimensional images generated using the MNIST dataset~\cite{lecun1998mnist}, CMNIST.
Specifically, the shape of $X$ is controlled by $Y$ while the color is determined by $Z$ (red for $Z=1$ and green for $Z=0$).
The shape is randomly sampled from digits in $\{5, 6, 7, 8, 9\}$ when $Y=1$ and from $\{0, 1, 2, 3, 4\}$ when $Y=0$.

For both datasets, multiple sites with different $\beta$ coefficients are generated (see Table~\ref{tab:syn_data}).
We considered two additional setups: multiple training sites and a single training site.
As there are 2 types of data, 3 causal relations between $Y$ and $Z$, and 2 different training setups, there are 12 different sets of results in total.

In the CMNIST experiments, we have an additional baseline, $\text{ERM}^c$, which takes greyscale images ($X'$) as input and is trained with ERM.
Thus, $\text{ERM}^c$~\footnote{\label{fn:ermc}Note $\text{ERM}^c$ has access to privileged information} ignores the effect of $Z$ and consequently is invariant to the unstable correlation between $Y$ and $Z$.
\begin{figure}[t]
\centering
\begin{subfigure}[t]{.5\linewidth}
\caption{2-dim}
\centering
\includegraphics[width=\linewidth]{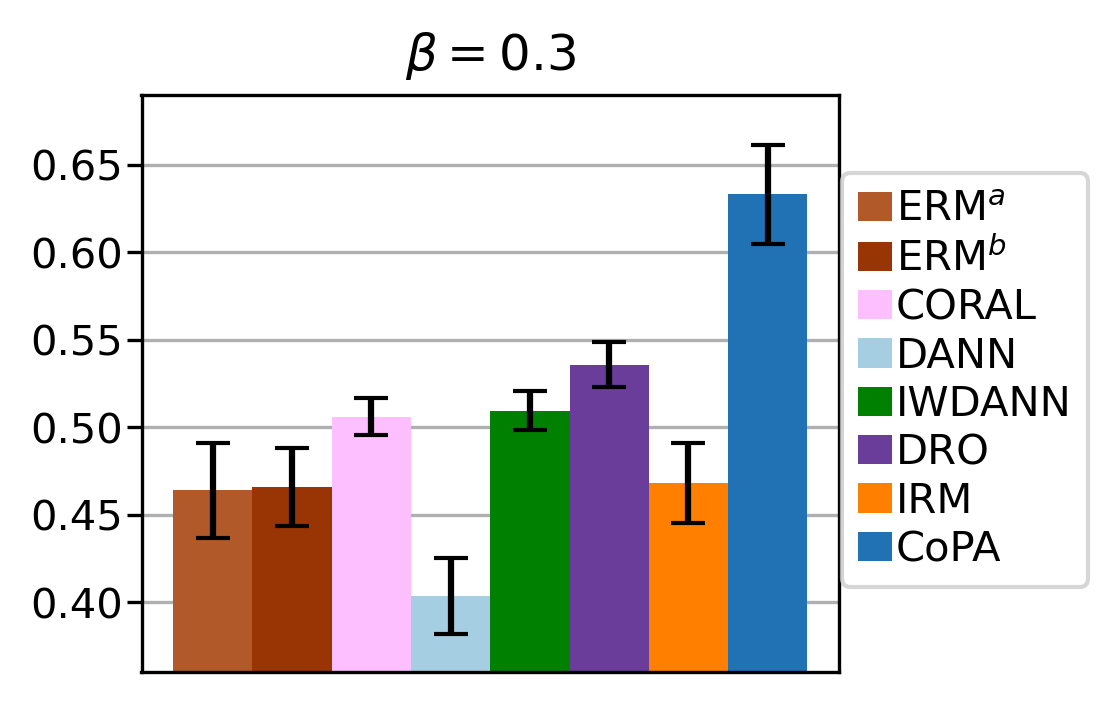}
\end{subfigure}%
\begin{subfigure}[t]{.5\linewidth}
\caption{CMNIST}\label{subfig:cmnist_YSZ}
\centering
\includegraphics[width=\linewidth]{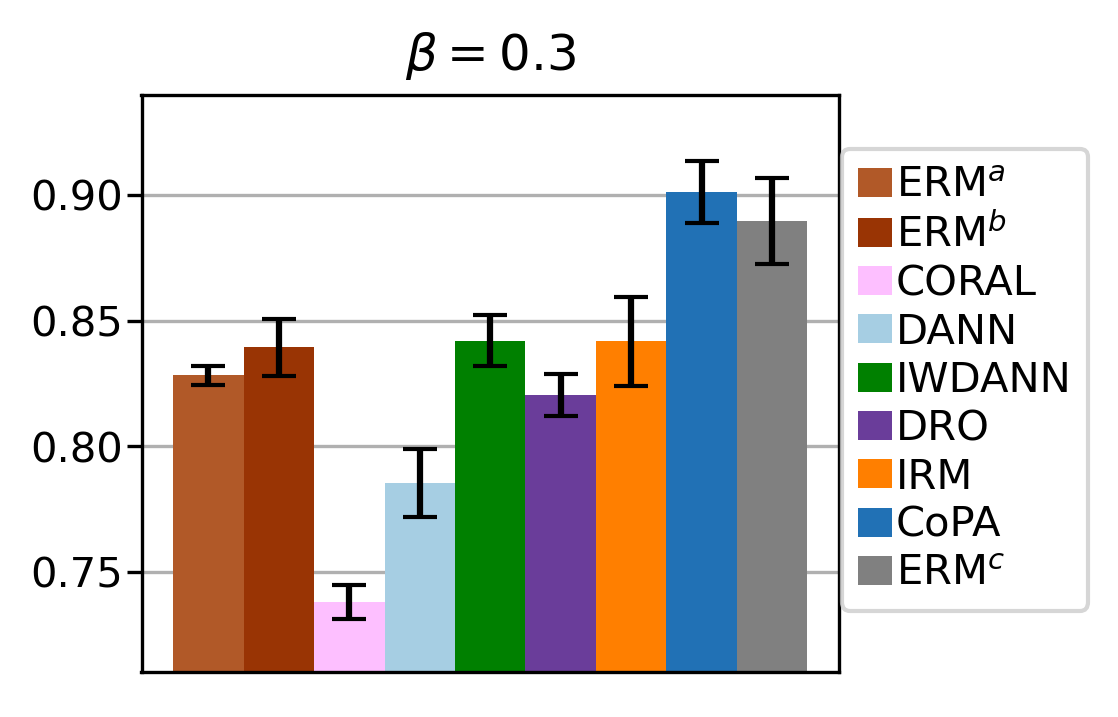}
\end{subfigure}
\caption{F1-score at test site, \emph{multiple} training sites.~\emph{$Y{\leftarrow}S{\rightarrow}Z$}
$\text{ERM}^a$: input=X,\;\; $\text{ERM}^b$: input=X,Z,\;\; $\text{ERM}^c$ \footref{fn:ermc}: greyscale input
}\label{fig:result_YSZ}
\end{figure}

\begin{figure}[t]
\centering
\begin{subfigure}[t]{.5\linewidth}
\caption{2-dim}
\centering
\includegraphics[width=\linewidth]{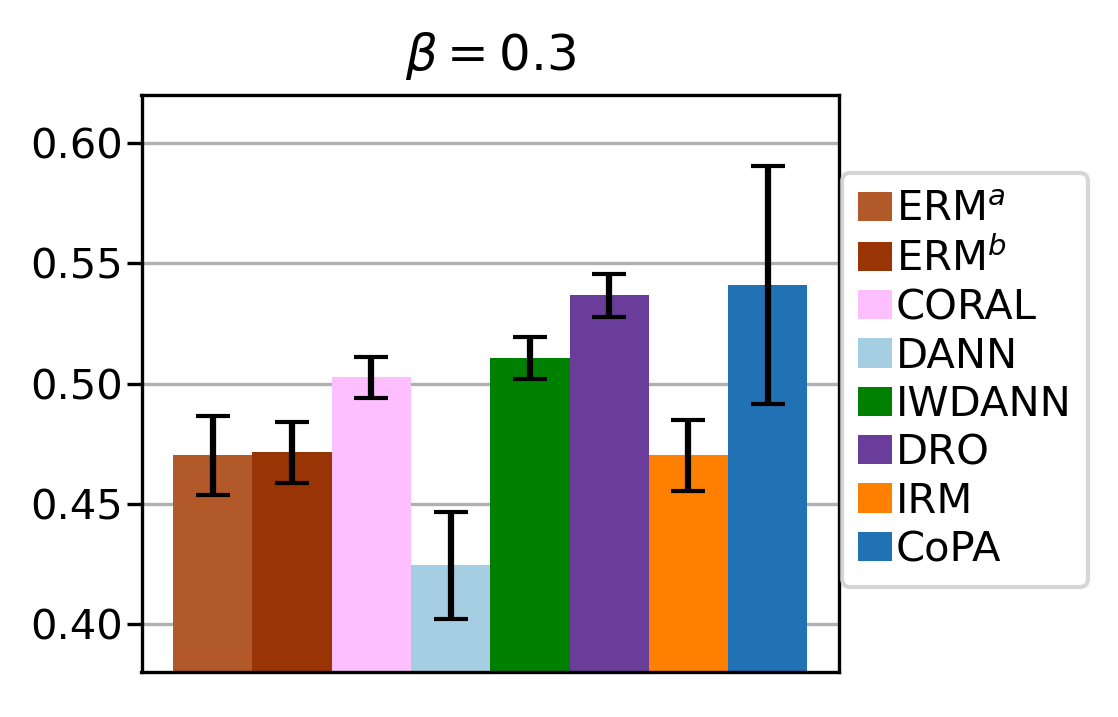}
\end{subfigure}%
\begin{subfigure}[t]{.5\linewidth}
\caption{CMNIST}\label{subfig:cmnist_Z}
\centering
\includegraphics[width=\linewidth]{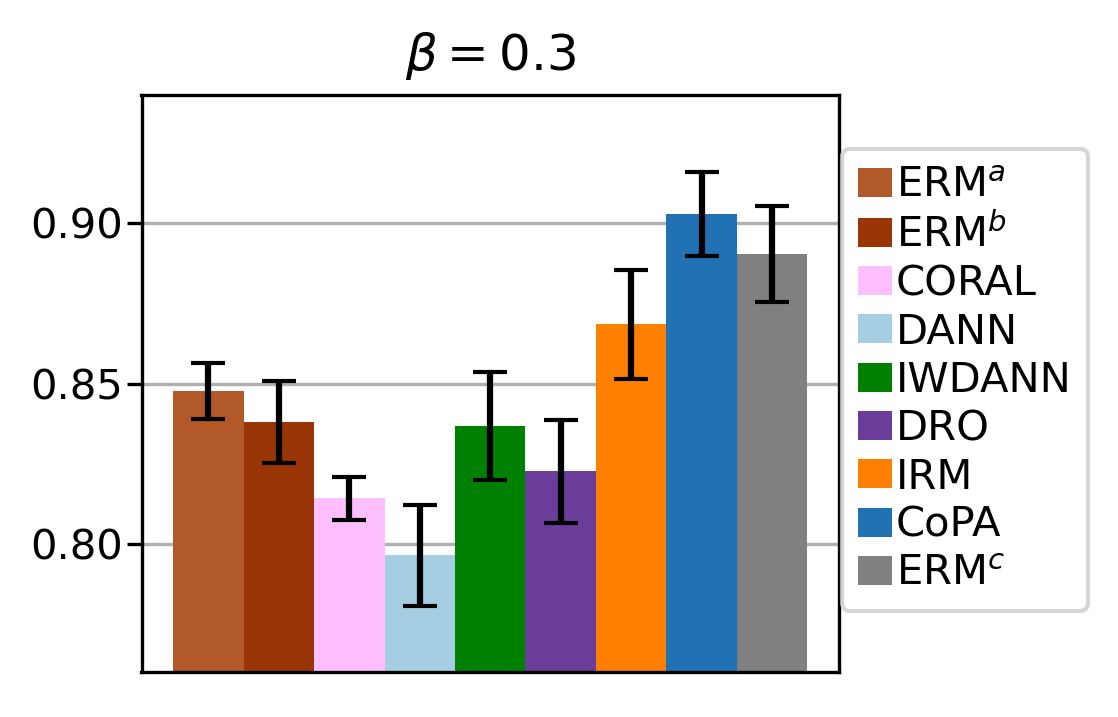}
\end{subfigure}
\caption{F1-score at test site, \emph{multiple} training sites.~\emph{$Z$ causes $Y$}.
$\text{ERM}^a$: input=X,\;\; $\text{ERM}^b$: input=X,Z,\;\; $\text{ERM}^c$ \footref{fn:ermc}: greyscale input
}\label{fig:result_Z}
\end{figure}
\subsubsection{Experimental Details}\label{sssec:syn_setup}
All compared methods used the same backbone network and were all trained with Adam~\cite{kingma2015adam} for 20k steps (convergence was confirmed by visual inspection) and 1e-4 learning rate.
For 2-dim data experiments, the backbone network was a single fully-connected (FC) layer with output dimension equal to 10.
For CMNIST data experiments, the backbone network was a CNN with three convolutional layers, each followed by $2 \times 2$ max-pooling and ReLU activation.
The numbers of channels and kernel size of the CNN layers were 32, 32, 64 and  $5\times 5$, $3\times 3$, $3\times 3$ respectively.
The output of the last convolutional layer is then flattened and fed through a FC layer with output dimension 256.

\begin{figure}[t]
\centering
\begin{subfigure}[t]{.5\linewidth}
\caption{2-dim}
\centering
\includegraphics[width=\linewidth]{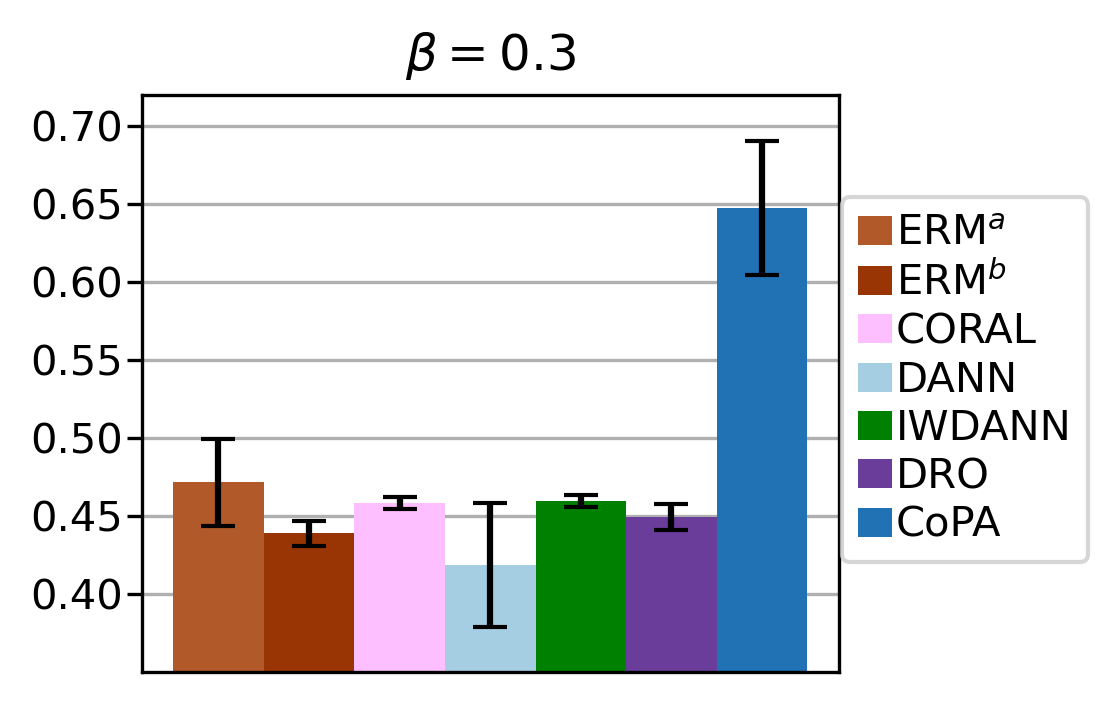}
\end{subfigure}%
\begin{subfigure}[t]{.5\linewidth}
\caption{CMNIST}
\centering
\includegraphics[width=\linewidth]{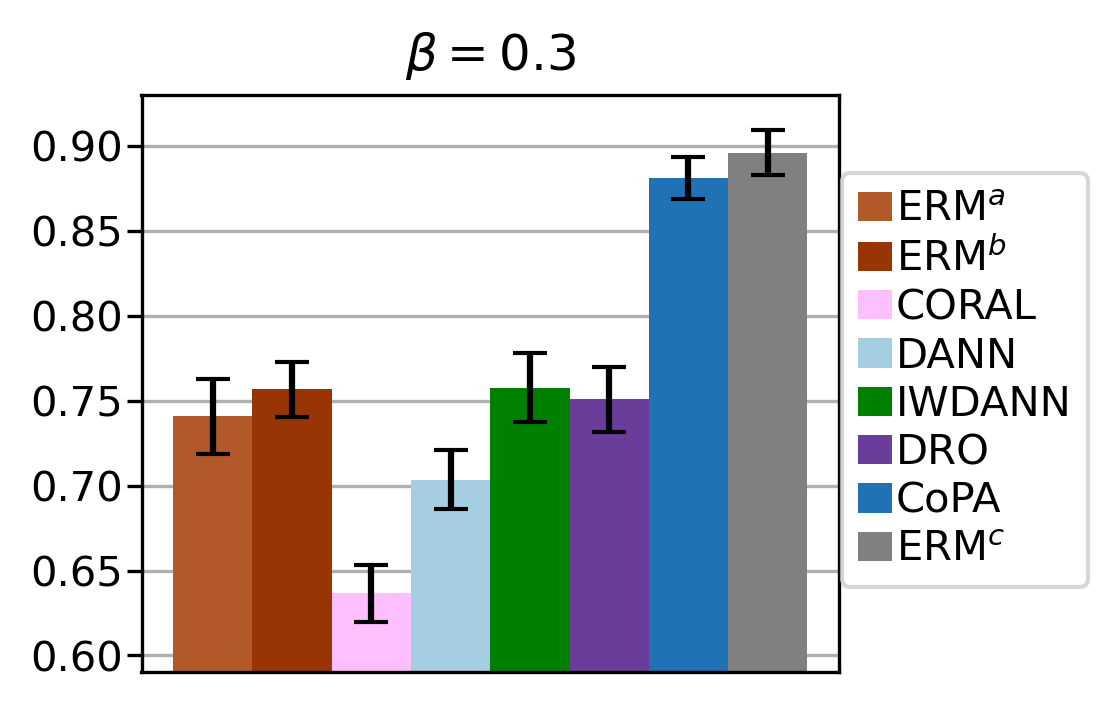}
\end{subfigure}
\caption{F1-score at test site, a \emph{single} training site.~\emph{$Y{\leftarrow}S{\rightarrow}Z$}
$\text{ERM}^a$: input=X,\;\; $\text{ERM}^b$: input=X,Z,\;\; $\text{ERM}^c$ \footref{fn:ermc}: greyscale input
}\label{fig:result_YSZ_112}
\end{figure}

\subsubsection{Results}\label{sssec:syn_result}
Figure~\ref{fig:result_YSZ} shows the test site performance when there are multiple training sites and $Y$ and $Z$ are spuriously correlated.
The lower the test site's $\beta$, the weaker the correlation between $Y$ and $Z$ and the stronger the label-shift.
When $\beta=0.3$, $Y$ and $Z$ are almost uncorrelated.
In this setup, \MNAME~outperforms all the baselines.
Note that there are 3 variants of ERM, each receiving a different input.
There is no consistent difference in performance between $\text{ERM}^a$ (only $X$ as input) and $\text{ERM}^b$ ($X$ and $Z$ as input).
In general, the other baselines do not consistently outperform ERM.
Although IWDANN outperforms DANN because the former also models label-shift, its performance is always worse than \MNAME.
In contrast, \MNAME~outperforms all baselines, including $\text{ERM}^c$ in Figure~\ref{subfig:cmnist_YSZ}.
This is because \MNAME~accounts for label-shift, while $\text{ERM}^c$ does not.
In addition, ignoring $Z$ may harm performance in the case when $Z$ is a cause of $Y$ (Figure~\ref{subfig:cmnist_Z}).
When there is only one training site (Figure~\ref{fig:result_YSZ_112}), \MNAME~is still better than baselines.

\begin{table*}[t]
\setlength{\tabcolsep}{4pt}
\centering
\begin{tabular}{lrrrrrrrrrr}
    Site ($E$)    &  BCN1 &  BCN2 &    MA &   NY1 &   \underline{NY2} &   \textbf{NY3} &   QLD &   \textbf{SYD} &  WIE1 &  WIE2 \\
    \midrule
    No.~of samples  &  7063 &  7311 &  9251 & 11108 &  1814 &  3186 &  8449 &  1884 &  7818 &  4374 \\
    \midrule
    Marginal prevalence, i.e~$P(Y{=}1|E)$ & 0.404 & 0.024 & 0.000 & 0.019 & 0.146 & 0.208 & 0.001 & 0.071 & 0.142 & 0.009
\end{tabular}
\caption{Different sites in ISIC.
The effect of label-shift (change in $P(Y|E)$) is very pronounced between sites.
\underline{Underlined}: validation site, \textbf{bolded}: test sites.
BCN: Barcelona, MA: Massachusetts, NY: New York, QLD: Queensland, SYD: Sydney, WIE: Vienna}\label{tab:isic_site}
\end{table*}
\subsection{Real Data Experiments}\label{ssec:real_exp}

\subsubsection{ISIC Data}\label{sssec:isic_data}
\begin{figure}[]
\centering
\includegraphics[width=\linewidth]{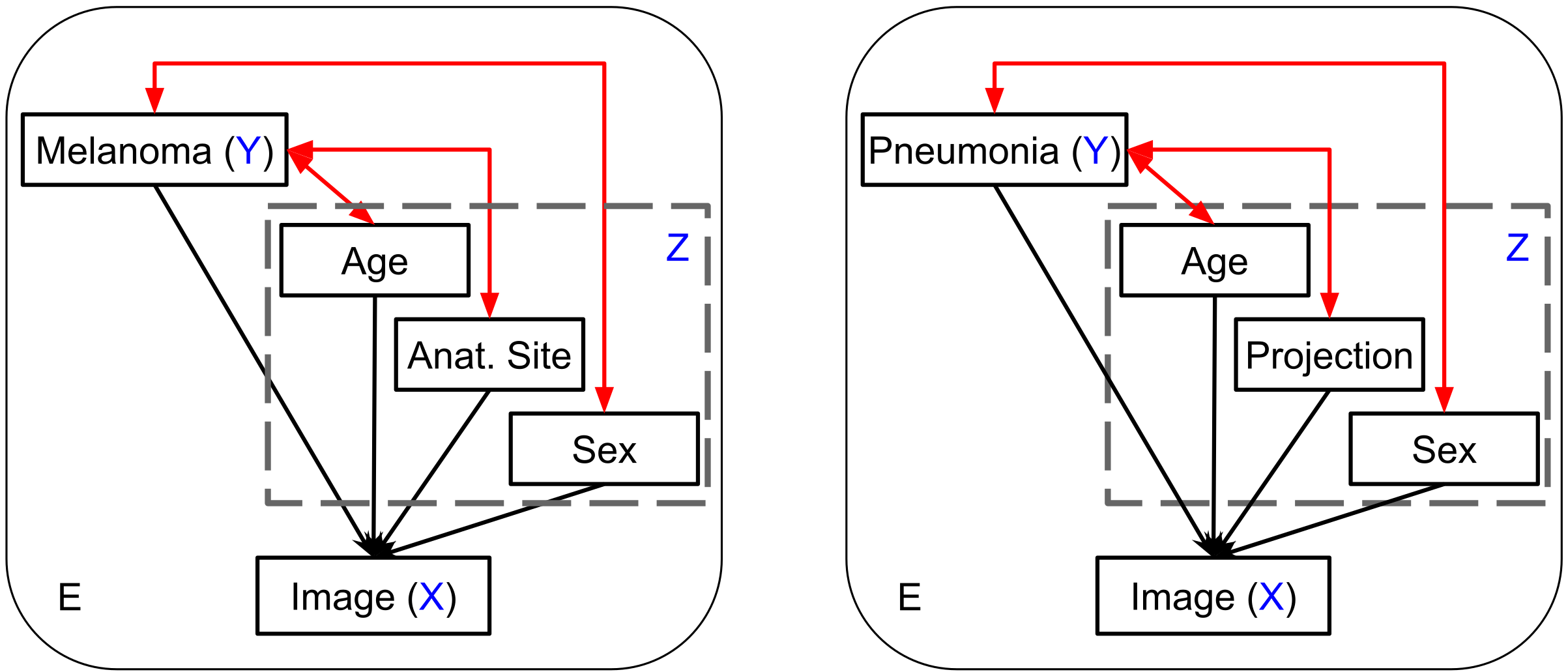}
\caption{Hypothesized causal graphs. (Left) ISIC (Right) CXR.
Bidirectional arrows indicate uncertainty in causal relationship.
}\label{fig:real_data_setup}
\end{figure}

The skin cancer dataset is from the International Skin Imaging Collaboration (ISIC) archive\footnote{https://www.isic-archive.com}.
Data from the archive~\cite{scope2009dermoscopic,gutman2016skin,codella2018skin,tschandl2018ham10000,codella2019skin,combalia2019bcn20000,rotemberg2021patient} are collected by different organizations at different points in time.
There are about 70k data samples in total (see Appendix~\ref{app:data_example} for some examples).
Each data sample consists of an input image $X$, a binary target label $Y$ (melanoma or not) and confounding variables $Z$ that is correlated with $Y$.
We consider three $Z$ variables: (1) \textit{Age}, (2) \textit{Anatomical Site} (there are 8 different sites, listed in Appendix~\ref{app:data_example}), and (3) \textit{Sex}.
While \textit{Age} is arguably a possible cause of $Y$~\cite{paulson2020age}, \textit{Anatomical Site} may be spuriously correlated with $Y$~\cite{lian2017natural} (Figure~\ref{fig:real_data_setup}, left panel).
The values of \textit{Age} in ISIC are discretized so \textit{Age} is a categorical variable.
Samples are grouped into sites based on spatio-temporal information as shown in Table~\ref{tab:isic_site}.
Table~\ref{tab:isic_site} also shows that the marginal prevalence of melanoma, $P(Y=1|E)$, varies drastically between sites.
Data from \textit{NY2} site were used for validation while data from \textit{NY3} and \textit{SYD} sites were used for testing.
The remaining sites were used for training.

\begin{table}[t]
\setlength{\tabcolsep}{4pt}
\centering
\begin{tabular}{lrrr}
    Site ($E$)    &  CXR8 &  \underline{CheXpert} &   \textbf{PadChest} \\
    \midrule
    No.~of samples  &  26202 &  5886 &  4592 \\
    \midrule
    $P(Y{=}1|E)$ & 0.049 & 0.635 & 0.082
\end{tabular}
\caption{Different sites and corresponding marginal prevalence ($P(Y|E)$) in CXR.
\underline{Underlined}: validation site, \textbf{bolded}: test site.}\label{tab:cxr_site}
\end{table}
\subsubsection{Chest X-Ray (CXR) Data}\label{sssec:cxr_data}
The Chest X-Ray data come from 3 datasets: CXR8~\cite{wang2017chestx}, CheXpert~\cite{irvin2019chexpert}, and PadChest~\cite{bustos2020padchest}.
Each data sample consists of an input image $X$, a binary target label $Y$ (having pneumonia or not) and confounding variables $Z$.
For CXR8 and PadChest~\cite{bustos2020padchest}, samples with ``No Finding'' label are used as negative target ($Y=0$)
We again consider three $Z$ variables: (1) \textit{Age}, (2) \textit{Projection} (AP, PA, or LL), and (3) \textit{Sex}.
Unlike ISIC, \textit{Age} is a continuous variable.
Table~\ref{tab:cxr_site} shows the training/validation/test sites and their corresponding marginal prevalence, $P(Y=1|E)$.

\begin{figure}[t]
\centering
\includegraphics[width=\linewidth]{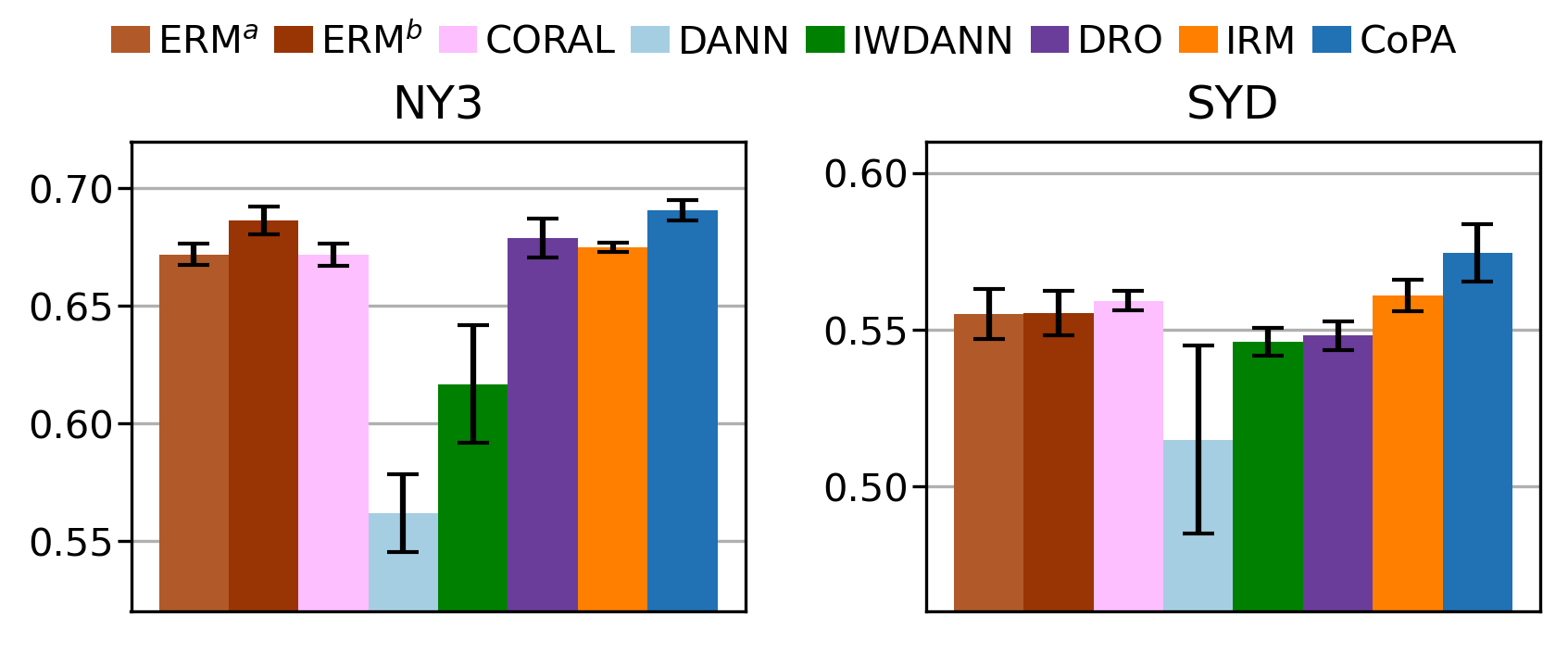}
\caption{F1-score at ISIC test sites, \emph{multiple} training sites.}\label{fig:isic_site}
\end{figure}
\begin{figure}[t]
\centering
\includegraphics[width=.6\linewidth]{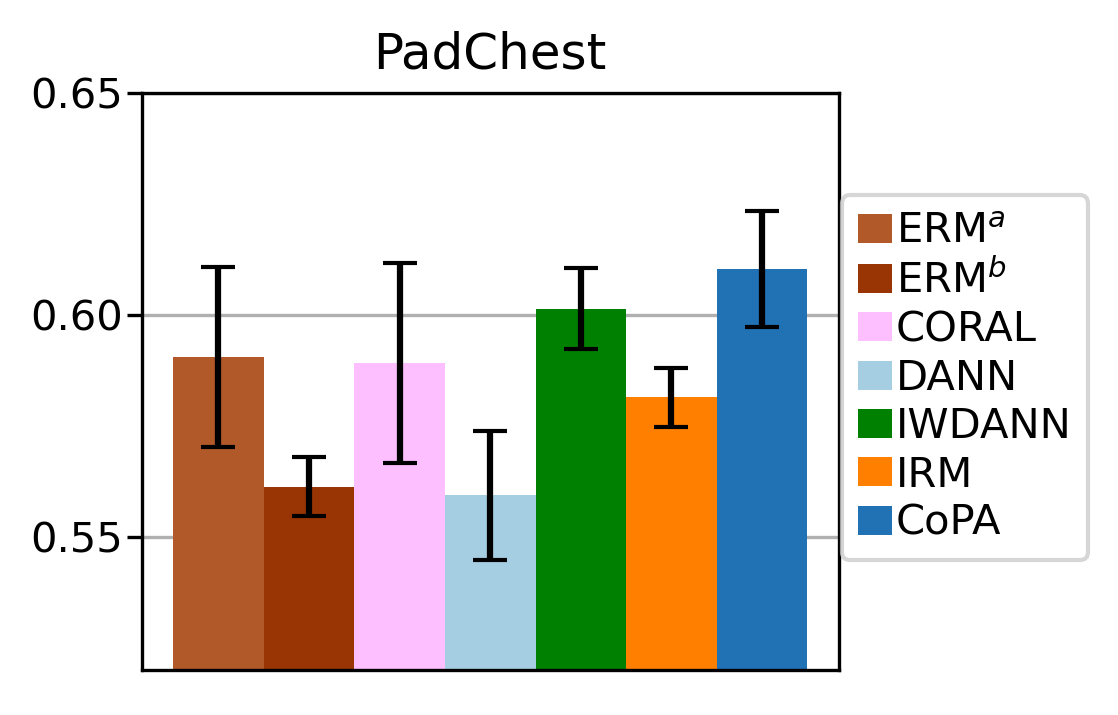}
\caption{F1-score at CXR test site, a \emph{single} training site}\label{fig:cxr_result}
\end{figure}
\subsubsection{Experimental Details}
All methods used a pre-trained ResNet50~\cite{he2016deep,vryniotis2021train} as the backbone.
ResNet50's output is then fed through an FC layer with output dimension 256.
Finetuning was done using Adam~\cite{kingma2015adam} for 20k steps with 3e-5 learning rate.
The site-specific prevalences are estimated by fitting a simple neural networks with 3 hidden layers with 20 hidden units each and ReLU activation.
The multiple variables $Z$ are concatenated together when used as input for \MNAME.
In ISIC setup, each combination of $Z$ is a group in DRO. In CXR setup, DRO is omitted since \textit{Age} in $Z$ is a continuous variable so there are infinitely many groups.

\subsubsection{Results}
For ISIC experiment, \MNAME~outperforms baseline methods at both \textit{NY3} and \textit{SYD} test sites (Figure~\ref{fig:isic_site}).
This also shows the flexibility of \MNAME, which can be applied to both sites with high prevalence, e.g.~\textit{NY3}, and sites with low prevalence, e.g.~\textit{SYD}.
For CXR experiment, \MNAME~also outperforms the baselines (Figure~\ref{fig:cxr_result}), demonstrating \MNAME's ability to work when only a single training site is available.

\section{Ablation}\label{sec:ablate}
\begin{figure}[t]
\centering
\begin{subfigure}[t]{\linewidth}
\caption{2-dim}
\centering
\includegraphics[width=.7\linewidth]{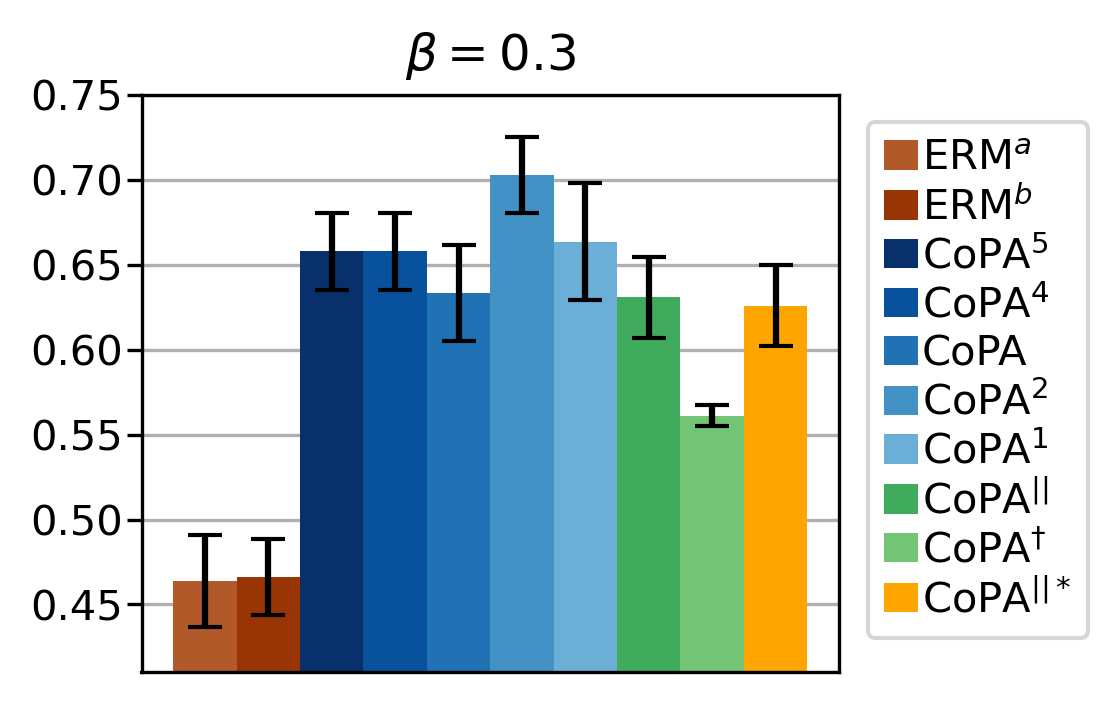}
\end{subfigure}
\begin{subfigure}[t]{\linewidth}
\caption{CMNIST}
\centering
\includegraphics[width=.7\linewidth]{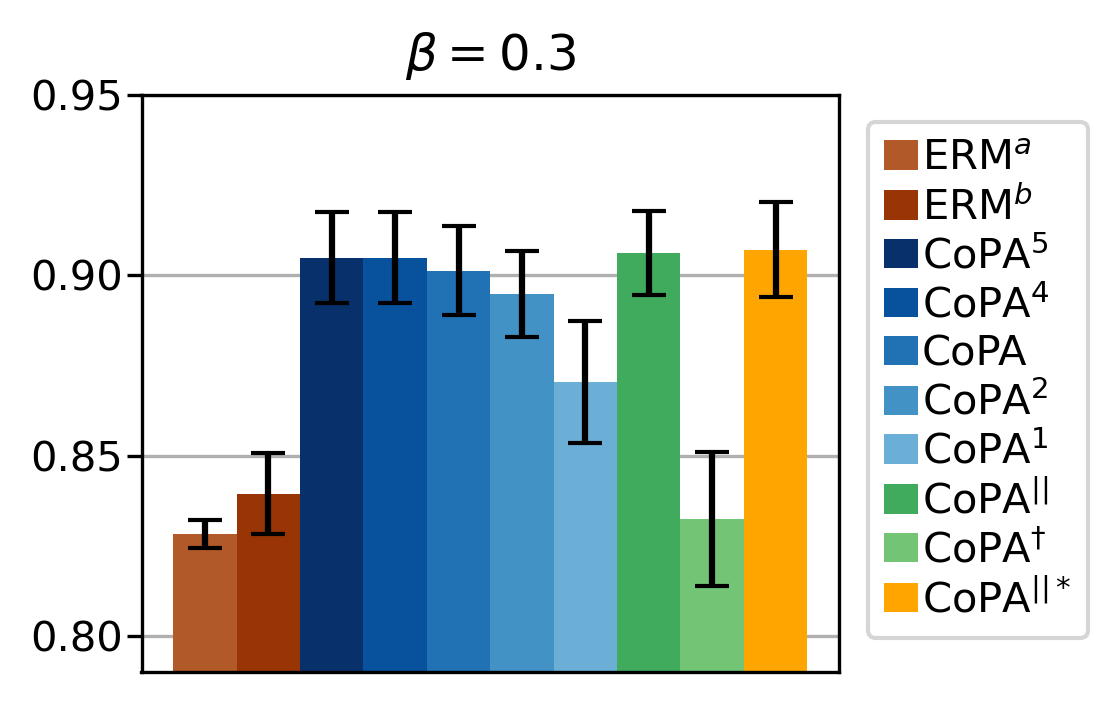}
\end{subfigure}
\caption{Ablation on synthetic data. Test F1-score. $Y{\leftarrow}S{\rightarrow}Z$
}\label{fig:ablate_YSZ}
\end{figure}

\begin{figure}[t]
\centering
\includegraphics[width=\linewidth]{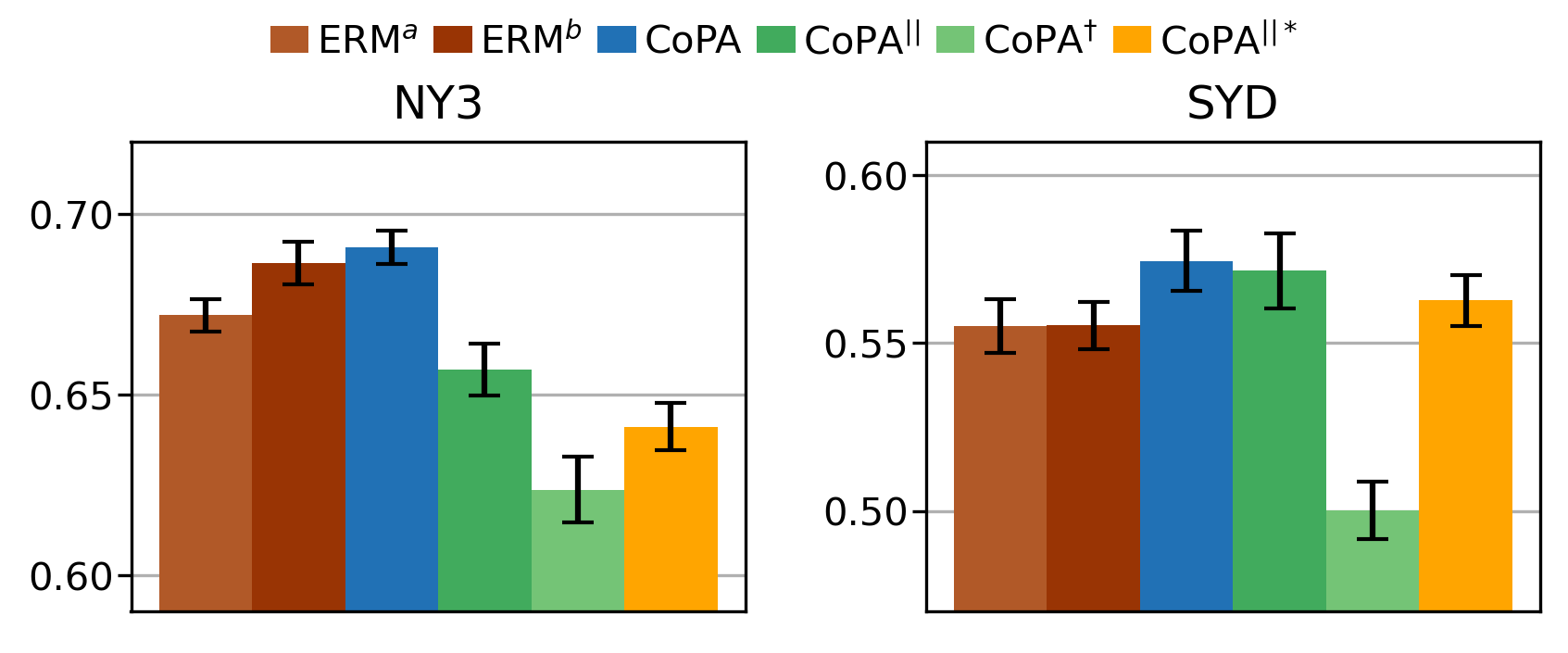}
\caption{Ablation on ISIC data. Test F1-score.
}\label{fig:ablate_isic}
\end{figure}

\begin{figure}[t]
\centering
\includegraphics[width=.7\linewidth]{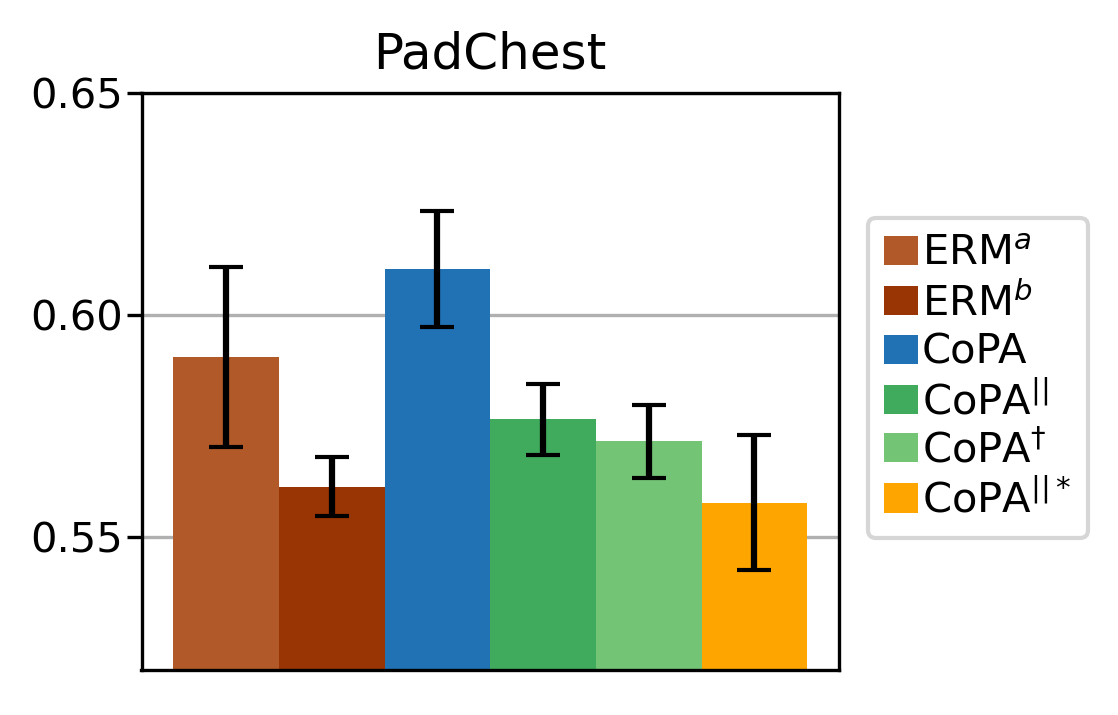}
\caption{Ablation on CXR data. Test F1-score.
}\label{fig:ablate_cxr}
\end{figure}
\MNAME~assumes the availability of the conditional prevalence $P(Y|Z,E)$ at each site $E$; and the observability of confounders $Z$ at training and test sites.
We examine how \MNAME's performance varies with less accurate prevalence estimates (former) and how \MNAME~can be used when confounders $Z$ are not observed at test sites (latter).

\subsection{Ablated Variants}\label{ssec:ablation_variants}
We analyzed how sensitive \MNAME~is to the accuracy of $\widehat{P}(Y|Z,e_i)$.
For synthetic data experiments, while keeping training unchanged, we varied $L_i$, the number of $(Y,Z)$ pairs used to estimate $\widehat{P}(Y|Z,e_i)$, at test sites.
The lower $L_i$ is, the less accurate $\widehat{P}(Y|Z,e_i)$.
Beside $L_i{=}10^3$ (denoted as \MNAME), we tested $L_i\in\{10^5,10^4,10^2,10\}$ (denoted as $\text{\MNAME}^5,\text{\MNAME}^4,\text{\MNAME}^2,\text{\MNAME}^1$ respectively).
We also tested: (1) the marginal prevalence $\widehat{P}(Y|e_i)$ (i.e.~$\text{\MNAME}^{\|}$) and (2) the uniform prevalence (i.e.~$\text{\MNAME}^{\dagger}$).
These estimates are even less accurate but are easier to obtain.
$\widehat{P}(Y|e_i)$ can replace $\widehat{P}(Y|Z,e_i)$ with no loss in performance if $Y\independent Z|E=e_i$.

When $Z$ is unknown, one can predict using the approximation $P(Y|X,e_i){=}\sum_{Z} P(Y|X,Z,e_i)$ and using $\widehat{P}(Y|e_i)$ instead of $\widehat{P}(Y|Z,e_i)$.
While this variant (i.e.~$\text{\MNAME}^{\|*}$) unrealistically assumes a uniform $P(Z|X,e_i)$, $\widehat{Y}$ may be correct despite the wrong probability estimate.
For high-dimensional $Z$, the summation is intractable so we implement a Monte Carlo strategy by summing over 10 random values of $Z$.

\subsection{Ablation Results}\label{ssec:ablation_results}
Figure~\ref{fig:ablate_YSZ} shows that the more accurate $\widehat{P}(Y|Z,e_i)$ is, the higher \MNAME's F1-score is in general.
Using the uniform prevalence ($\text{\MNAME}^{\dagger}$) is generally bad while using the marginal prevalence ($\text{\MNAME}^{\|}$) can be acceptable when $Y$ and $Z$ are uncorrelated ($\beta=0.3$).
Figure~\ref{fig:ablate_isic} \textit{SYD} also shows $\widehat{P}(Y|e_i)$ can be an acceptable substitute for $\widehat{P}(Y|Z,e_i)$.
Besides, it seems that $\text{\MNAME}^{\|*}$ occasionally outperforms ERM.

\section{Discussion}\label{sec:discussion}
In this work, we propose \MNAME: an approach for domain-invariant representation learning for anti-causal problems by adjusting for the effect of unstable correlations through the conditional prevalence estimate.
By learning a stable predictor of $Y$ that leverages the stable edges and an estimate of the prevalence in each site, \MNAME~can work regardless of (1) the number of training sites available, (2) the presence or absence of label-shift, and (3) a variable relationship between $Y$ and confounding $Z$ variable(s) (spurious or causal).
Our core insight is that in many applications it can be possible to infer the prevalence in each site, including the test site(s), as one only needs a set of $(Y,Z)$ samples.
Crucially, we assume $Z$'s are observed, but no labeled $X$'s are necessary for the test site.
Our experiments on synthetic datasets and two real medical imaging datasets show \MNAME~can outperform competitive baselines.
In particular, our ablation study demonstrates that \MNAME~can still be useful even if our prevalence estimate is naive or sub-optimal.

A core weakness of \MNAME~is that it assumes that confounding variable(s) $Z$ are observed, which is often the case in healthcare settings but might not be true in other applications.
Although, our ablation results show tolerable performance when $Z$ is not observed, more rigorous treatment of this case is warranted.

{\small
\bibliographystyle{ieee_fullname}
\bibliography{ref_cause,ref_crepr,ref_med}

\begin{thebibliography}{10}\itemsep=-1pt

\bibitem{lecun1998mnist}
The mnist database of handwritten digits.
\newblock {\em http://yann. lecun. com/exdb/mnist/}.

\bibitem{ahuja2021invariance}
Kartik Ahuja, Ethan Caballero, Dinghuai Zhang, Jean-Christophe Gagnon-Audet,
  Yoshua Bengio, Ioannis Mitliagkas, and Irina Rish.
\newblock Invariance principle meets information bottleneck for
  out-of-distribution generalization.
\newblock In {\em Proceedings of NeurIPS}, volume~34, pages 3438--3450, 2021.

\bibitem{albadawy2018deep}
Ehab~A AlBadawy, Ashirbani Saha, and Maciej~A Mazurowski.
\newblock Deep learning for segmentation of brain tumors: Impact of
  cross-institutional training and testing.
\newblock {\em Medical physics}, 45(3):1150--1158, 2018.

\bibitem{arjovsky2019invariant}
Martin Arjovsky, L{\'e}on Bottou, Ishaan Gulrajani, and David Lopez-Paz.
\newblock Invariant risk minimization.
\newblock Eprint \href{http://arxiv.org/abs/1907.02893}{arXiv:1907.02893},
  2019.

\bibitem{bustos2020padchest}
Aurelia Bustos, Antonio Pertusa, Jose-Maria Salinas, and Maria de~la
  Iglesia-Vay{\'a}.
\newblock Padchest: A large chest x-ray image dataset with multi-label
  annotated reports.
\newblock {\em Medical image analysis}, 66:101797, 2020.

\bibitem{codella2019skin}
Noel Codella, Veronica Rotemberg, Philipp Tschandl, M~Emre Celebi, Stephen
  Dusza, David Gutman, Brian Helba, Aadi Kalloo, Konstantinos Liopyris, Michael
  Marchetti, et~al.
\newblock Skin lesion analysis toward melanoma detection 2018: A challenge
  hosted by the international skin imaging collaboration (isic).
\newblock Eprint \href{http://arxiv.org/abs/1902.03368}{arXiv:1902.03368},
  2019.

\bibitem{codella2018skin}
Noel~CF Codella, David Gutman, M~Emre Celebi, Brian Helba, Michael~A Marchetti,
  Stephen~W Dusza, Aadi Kalloo, Konstantinos Liopyris, Nabin Mishra, Harald
  Kittler, et~al.
\newblock Skin lesion analysis toward melanoma detection: A challenge at the
  2017 international symposium on biomedical imaging (isbi), hosted by the
  international skin imaging collaboration (isic).
\newblock In {\em Proceedings of ISBI}, pages 168--172. IEEE, 2018.

\bibitem{combalia2019bcn20000}
Marc Combalia, Noel~CF Codella, Veronica Rotemberg, Brian Helba, Veronica
  Vilaplana, Ofer Reiter, Cristina Carrera, Alicia Barreiro, Allan~C Halpern,
  Susana Puig, et~al.
\newblock Bcn20000: Dermoscopic lesions in the wild.
\newblock Eprint \href{http://arxiv.org/abs/1908.02288}{arXiv:1908.02288},
  2019.

\bibitem{degrave2021ai}
Alex~J DeGrave, Joseph~D Janizek, and Su-In Lee.
\newblock Ai for radiographic covid-19 detection selects shortcuts over signal.
\newblock {\em Nature Machine Intelligence}, 3(7):610--619, 2021.

\bibitem{ganin2016domain}
Yaroslav Ganin, Evgeniya Ustinova, Hana Ajakan, Pascal Germain, Hugo
  Larochelle, Fran{\c{c}}ois Laviolette, Mario Marchand, and Victor Lempitsky.
\newblock Domain-adversarial training of neural networks.
\newblock {\em JMLR}, 17(1):2096--2030, 2016.

\bibitem{gulrajani2021search}
Ishaan Gulrajani and David Lopez-Paz.
\newblock In search of lost domain generalization.
\newblock In {\em Proceedings of ICLR}, 2021.

\bibitem{gutman2016skin}
David Gutman, Noel~CF Codella, Emre Celebi, Brian Helba, Michael Marchetti,
  Nabin Mishra, and Allan Halpern.
\newblock Skin lesion analysis toward melanoma detection: A challenge at the
  international symposium on biomedical imaging (isbi) 2016, hosted by the
  international skin imaging collaboration (isic).
\newblock Eprint \href{http://arxiv.org/abs/1605.01397}{arXiv:1605.01397},
  2016.

\bibitem{he2016deep}
Kaiming He, Xiangyu Zhang, Shaoqing Ren, and Jian Sun.
\newblock Deep residual learning for image recognition.
\newblock In {\em Proceedings of CVPR}, pages 770--778, 2016.

\bibitem{irvin2019chexpert}
Jeremy Irvin, Pranav Rajpurkar, Michael Ko, Yifan Yu, Silviana Ciurea-Ilcus,
  Chris Chute, Henrik Marklund, Behzad Haghgoo, Robyn Ball, Katie Shpanskaya,
  et~al.
\newblock Chexpert: A large chest radiograph dataset with uncertainty labels
  and expert comparison.
\newblock In {\em Proceedings of AAAI}, volume~33, pages 590--597, 2019.

\bibitem{kamath2021does}
Pritish Kamath, Akilesh Tangella, Danica Sutherland, and Nathan Srebro.
\newblock Does invariant risk minimization capture invariance?
\newblock In {\em Proceedings of AISTATS}, pages 4069--4077. PMLR, 2021.

\bibitem{kingma2015adam}
Diederik~P. Kingma and Jimmy Ba.
\newblock Adam: {A} {Method} for {Stochastic} {Optimization}.
\newblock In {\em Proceedings of ICLR}, 2014.

\bibitem{koh2021wilds}
Pang~Wei Koh, Shiori Sagawa, Henrik Marklund, Sang~Michael Xie, Marvin Zhang,
  Akshay Balsubramani, Weihua Hu, Michihiro Yasunaga, Richard~Lanas Phillips,
  Irena Gao, et~al.
\newblock Wilds: A benchmark of in-the-wild distribution shifts.
\newblock In {\em Proceedings of ICML}, pages 5637--5664. PMLR, 2021.

\bibitem{krueger2021out}
David Krueger, Ethan Caballero, Joern-Henrik Jacobsen, Amy Zhang, Jonathan
  Binas, Dinghuai Zhang, Remi Le~Priol, and Aaron Courville.
\newblock Out-of-distribution generalization via risk extrapolation (rex).
\newblock In {\em Proceedings of ICML}, pages 5815--5826. PMLR, 2021.

\bibitem{le2021lamda}
Trung Le, Tuan Nguyen, Nhat Ho, Hung Bui, and Dinh Phung.
\newblock Lamda: Label matching deep domain adaptation.
\newblock In {\em Proceedings of ICML}, pages 6043--6054. PMLR, 2021.

\bibitem{li2018domain}
Ya Li, Mingming Gong, Xinmei Tian, Tongliang Liu, and Dacheng Tao.
\newblock Domain generalization via conditional invariant representations.
\newblock In {\em Proceedings of AAAI}, volume~32, 2018.

\bibitem{li2018deep}
Ya Li, Xinmei Tian, Mingming Gong, Yajing Liu, Tongliang Liu, Kun Zhang, and
  Dacheng Tao.
\newblock Deep domain generalization via conditional invariant adversarial
  networks.
\newblock In {\em Proceedings of ECCV}, pages 624--639, 2018.

\bibitem{lian2017natural}
B Lian, CL Cui, L Zhou, X Song, XS Zhang, D Wu, L Si, ZH Chi, XN Sheng, LL Mao,
  et~al.
\newblock The natural history and patterns of metastases from mucosal melanoma:
  an analysis of 706 prospectively-followed patients.
\newblock {\em Annals of Oncology}, 28(4):868--873, 2017.

\bibitem{lipton2018detecting}
Zachary Lipton, Yu-Xiang Wang, and Alexander Smola.
\newblock Detecting and correcting for label shift with black box predictors.
\newblock In {\em Proceedings of ICML}, pages 3122--3130. PMLR, 2018.

\bibitem{muandet2013domain}
Krikamol Muandet, David Balduzzi, and Bernhard Sch{\"o}lkopf.
\newblock Domain generalization via invariant feature representation.
\newblock In {\em Proceedings of ICML}, pages 10--18. PMLR, 2013.

\bibitem{pan2010survey}
Sinno~Jialin Pan and Qiang Yang.
\newblock A survey on transfer learning.
\newblock {\em IEEE Transactions on knowledge and data engineering},
  22(10):1345--1359, 2010.

\bibitem{paulson2020age}
Kelly~G Paulson, Deepti Gupta, Teresa~S Kim, Joshua~R Veatch, David~R Byrd,
  Shailender Bhatia, Katherine Wojcik, Aude~G Chapuis, John~A Thompson,
  Margaret~M Madeleine, et~al.
\newblock Age-specific incidence of melanoma in the united states.
\newblock {\em JAMA dermatology}, 156(1):57--64, 2020.

\bibitem{peters2016causal}
Jonas Peters, Peter B{\"u}hlmann, and Nicolai Meinshausen.
\newblock Causal inference by using invariant prediction: identification and
  confidence intervals.
\newblock {\em Journal of the Royal Statistical Society: Series B (Statistical
  Methodology)}, 78(5):947--1012, 2016.

\bibitem{pooch2020can}
Eduardo~HP Pooch, Pedro Ballester, and Rodrigo~C Barros.
\newblock Can we trust deep learning based diagnosis? the impact of domain
  shift in chest radiograph classification.
\newblock In {\em Thoracic Image Analysis Workshop}, pages 74--83. Springer,
  2020.

\bibitem{rosenfeld2021risks}
Elan Rosenfeld, Pradeep~Kumar Ravikumar, and Andrej Risteski.
\newblock The risks of invariant risk minimization.
\newblock In {\em Proceedings of ICLR}, 2021.

\bibitem{rotemberg2021patient}
Veronica Rotemberg, Nicholas Kurtansky, Brigid Betz-Stablein, Liam Caffery,
  Emmanouil Chousakos, Noel Codella, Marc Combalia, Stephen Dusza, Pascale
  Guitera, David Gutman, et~al.
\newblock A patient-centric dataset of images and metadata for identifying
  melanomas using clinical context.
\newblock {\em Scientific data}, 8(1):1--8, 2021.

\bibitem{sagawa2019distributionally}
Shiori Sagawa, Pang~Wei Koh, Tatsunori~B Hashimoto, and Percy Liang.
\newblock Distributionally robust neural networks for group shifts: On the
  importance of regularization for worst-case generalization.
\newblock In {\em Proceedings of ICLR}, 2019.

\bibitem{scholkopf2012causal}
B Sch{\"o}lkopf, D Janzing, J Peters, E Sgouritsa, K Zhang, and J Mooij.
\newblock On causal and anticausal learning.
\newblock In {\em Proceedings of ICML}, pages 1255--1262, 2012.

\bibitem{schrouff2022maintaining}
Jessica Schrouff, Natalie Harris, Oluwasanmi Koyejo, Ibrahim Alabdulmohsin, Eva
  Schnider, Krista Opsahl-Ong, Alex Brown, Subhrajit Roy, Diana Mincu,
  Christina Chen, et~al.
\newblock Maintaining fairness across distribution shift: do we have viable
  solutions for real-world applications?
\newblock Eprint \href{http://arxiv.org/abs/2202.01034}{arXiv:2202.01034},
  2022.

\bibitem{scope2009dermoscopic}
A Scope, AA Marghoob, CS Chen, JA Lieb, MA Weinstock, AC Halpern, and
  SONIC~Study Group.
\newblock Dermoscopic patterns and subclinical melanocytic nests in
  normal-appearing skin.
\newblock {\em British Journal of Dermatology}, 160(6):1318--1321, 2009.

\bibitem{subbaswamy2022unifying}
Adarsh Subbaswamy, Bryant Chen, and Suchi Saria.
\newblock A unifying causal framework for analyzing dataset shift-stable
  learning algorithms.
\newblock {\em Journal of Causal Inference}, 10(1):64--89, 2022.

\bibitem{sun2016deep}
Baochen Sun and Kate Saenko.
\newblock Deep coral: Correlation alignment for deep domain adaptation.
\newblock In {\em Proceedings of ECCV}, pages 443--450. Springer, 2016.

\bibitem{tachet2020domain}
Remi Tachet~des Combes, Han Zhao, Yu-Xiang Wang, and Geoffrey~J Gordon.
\newblock Domain adaptation with conditional distribution matching and
  generalized label shift.
\newblock In {\em Proceedings of NeurIPS}, volume~33, pages 19276--19289, 2020.

\bibitem{tanwani2021dirl}
Ajay Tanwani.
\newblock Dirl: Domain-invariant representation learning for sim-to-real
  transfer.
\newblock In {\em Proceedings of CoRL}, pages 1558--1571. PMLR, 2021.

\bibitem{tschandl2018ham10000}
Philipp Tschandl, Cliff Rosendahl, and Harald Kittler.
\newblock The ham10000 dataset, a large collection of multi-source
  dermatoscopic images of common pigmented skin lesions.
\newblock {\em Scientific data}, 5(1):1--9, 2018.

\bibitem{vryniotis2021train}
Vasilis Vryniotis.
\newblock How to train state-of-the-art models using torchvision’s latest
  primitives, 2021.

\bibitem{wald2021calibration}
Yoav Wald, Amir Feder, Daniel Greenfeld, and Uri Shalit.
\newblock On calibration and out-of-domain generalization.
\newblock In {\em Proceedings of NeurIPS}, 2021.

\bibitem{wang2017chestx}
Xiaosong Wang, Yifan Peng, Le Lu, Zhiyong Lu, Mohammadhadi Bagheri, and
  Ronald~M Summers.
\newblock Chestx-ray8: Hospital-scale chest x-ray database and benchmarks on
  weakly-supervised classification and localization of common thorax diseases.
\newblock In {\em Proceedings of CVPR}, pages 2097--2106, 2017.

\bibitem{wang2022cisa}
Zihao Wang and Victor Veitch.
\newblock The causal structure of domain invariant supervised representation
  learning, 2022.

\bibitem{wilson2020survey}
Garrett Wilson and Diane~J Cook.
\newblock A survey of unsupervised deep domain adaptation.
\newblock {\em ACM Transactions on Intelligent Systems and Technology (TIST)},
  11(5):1--46, 2020.

\bibitem{zhao2019learning}
Han Zhao, Remi~Tachet Des~Combes, Kun Zhang, and Geoffrey Gordon.
\newblock On learning invariant representations for domain adaptation.
\newblock In {\em Proceedings of ICML}, pages 7523--7532. PMLR, 2019.

\end{thebibliography}
}

\clearpage
\appendix

\section{Complete Results}\label{app:complete_results}

\begin{figure}[ht]
\centering
\begin{subfigure}[t]{\linewidth}
\caption{2-dim,\;\; $\text{ERM}^a$: input=X,\;\; $\text{ERM}^b$: input=X,Z}
\centering
\includegraphics[width=.7\linewidth]{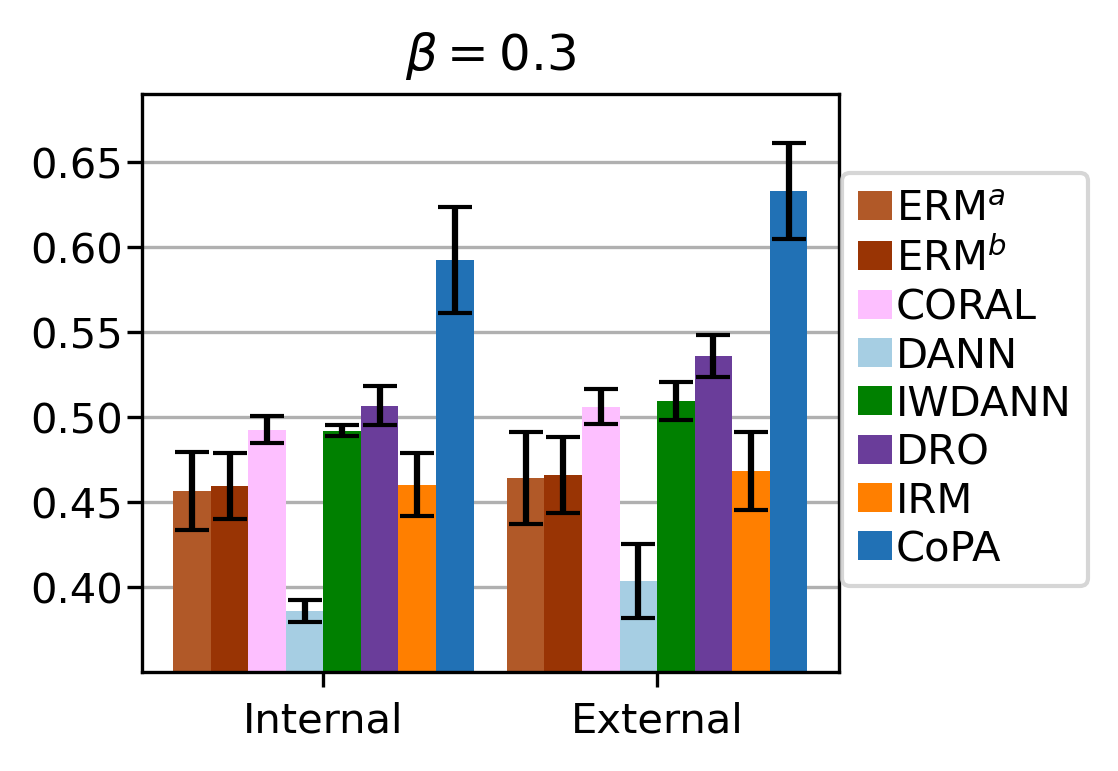}
\end{subfigure}
\begin{subfigure}[t]{\linewidth}
\caption{CMNIST,\;\; $\text{ERM}^c$: greyscale input}
\centering
\includegraphics[width=.7\linewidth]{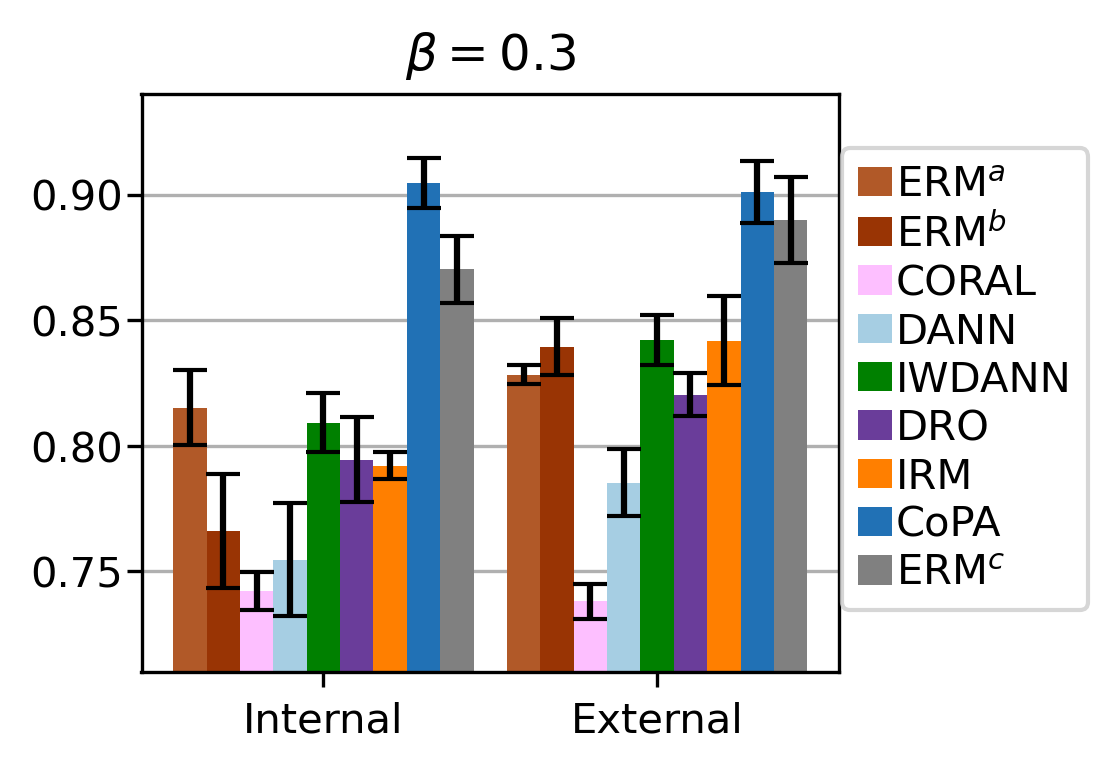}
\end{subfigure}
\caption{F1-score at test site, \emph{multiple} training sites.~\emph{$Y{\leftarrow}S{\rightarrow}Z$}
}\label{}
\end{figure}

\begin{figure}[ht]
\centering
\begin{subfigure}[t]{\linewidth}
\caption{2-dim,\;\; $\text{ERM}^a$: input=X,\;\; $\text{ERM}^b$: input=X,Z}
\centering
\includegraphics[width=.7\linewidth]{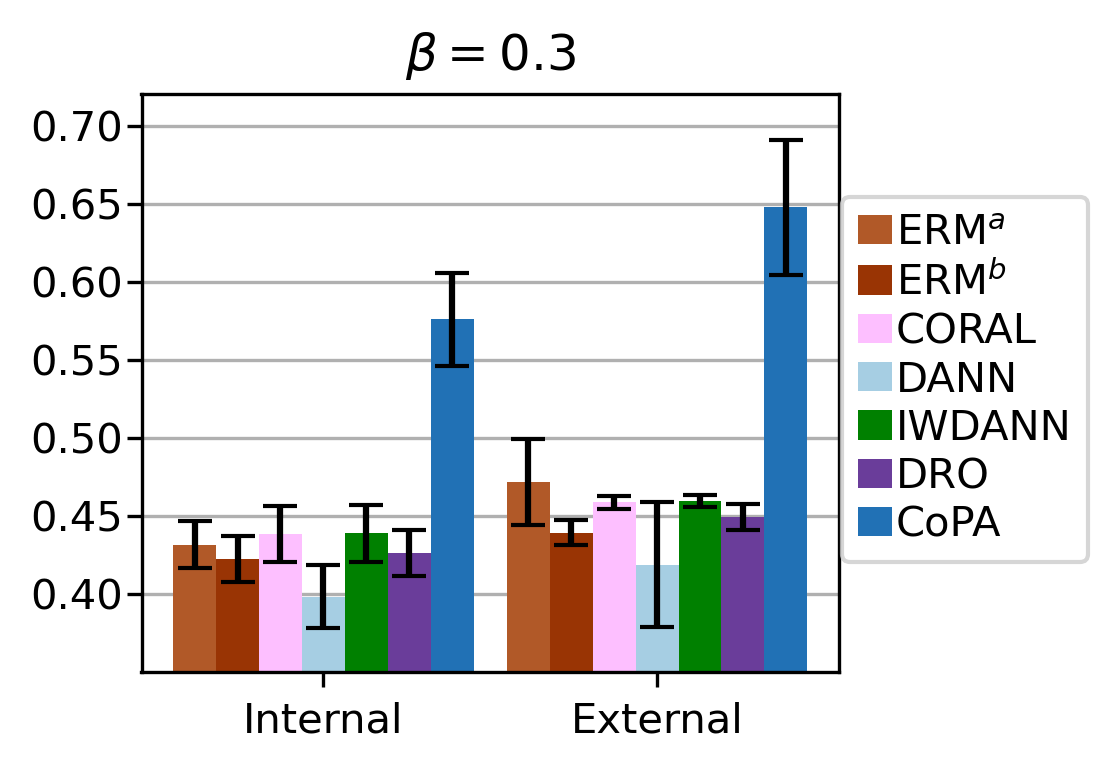}
\end{subfigure}
\begin{subfigure}[t]{\linewidth}
\caption{CMNIST,\;\; $\text{ERM}^c$: greyscale input}
\centering
\includegraphics[width=.7\linewidth]{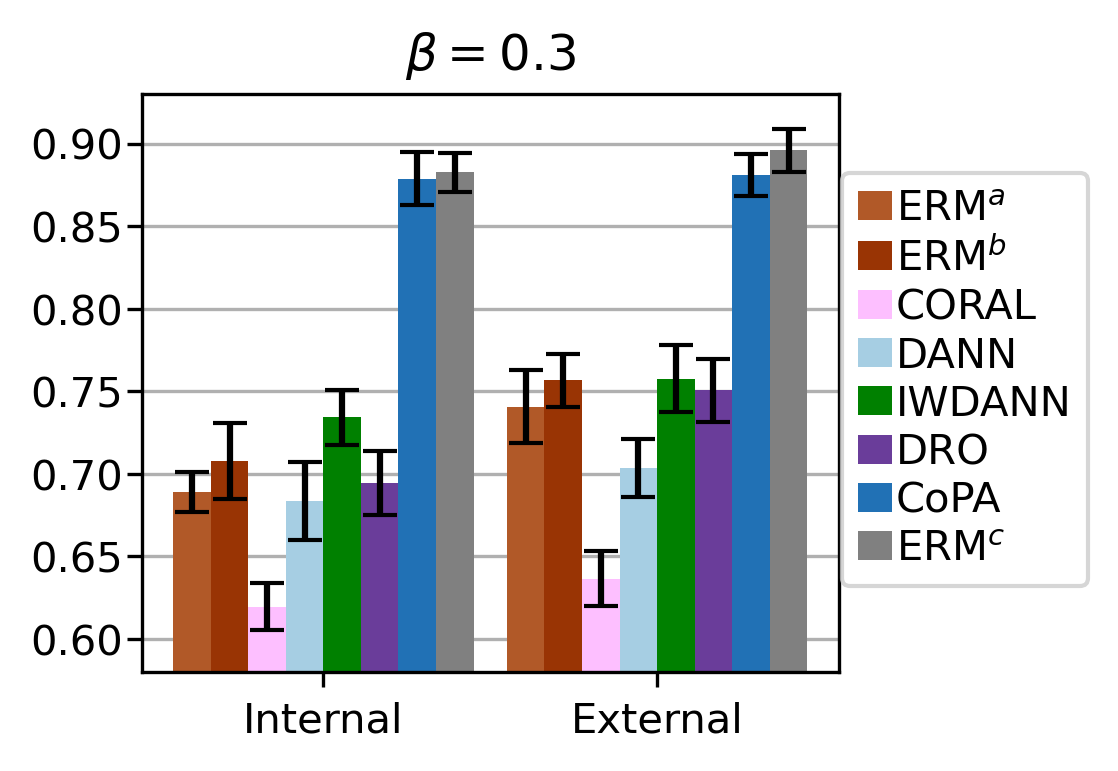}
\end{subfigure}
\caption{F1-score at test site, a \emph{single} training site.~\emph{$Y{\leftarrow}S{\rightarrow}Z$}
}\label{}
\end{figure}

\begin{figure}[ht]
\centering
\begin{subfigure}[t]{\linewidth}
\caption{2-dim,\;\; $\text{ERM}^a$: input=X,\;\; $\text{ERM}^b$: input=X,Z}
\centering
\includegraphics[width=.7\linewidth]{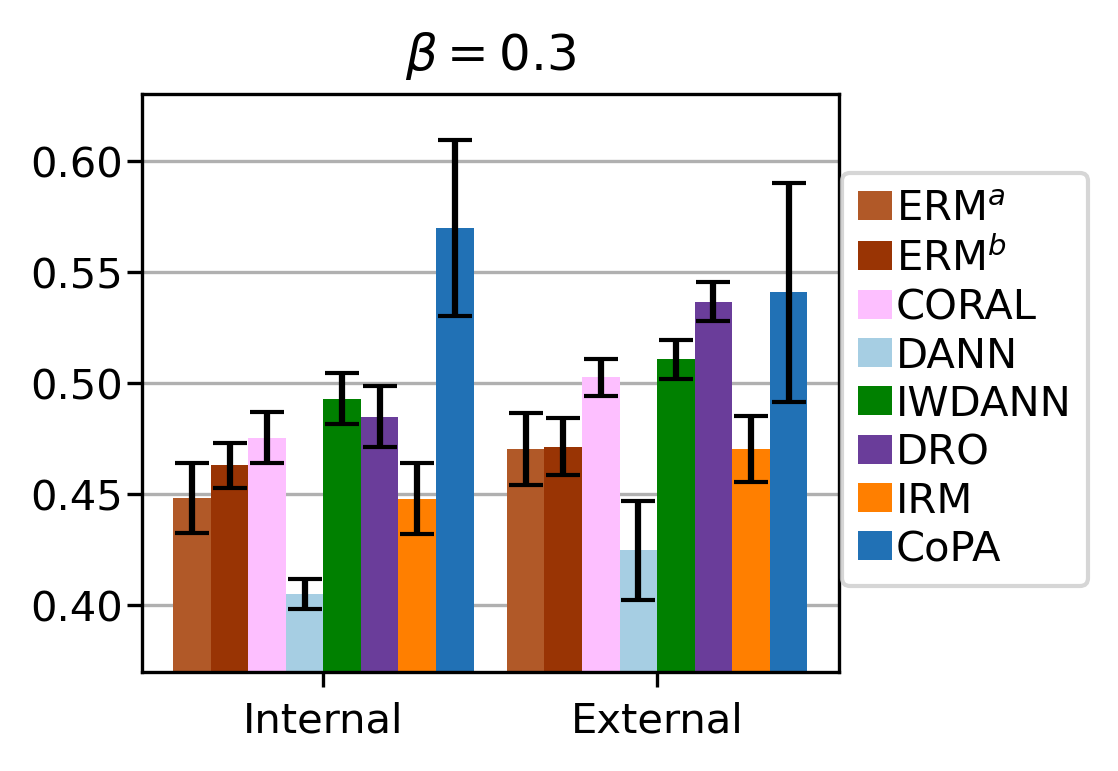}
\end{subfigure}
\begin{subfigure}[t]{\linewidth}
\caption{CMNIST,\;\; $\text{ERM}^c$: greyscale input}
\centering
\includegraphics[width=.7\linewidth]{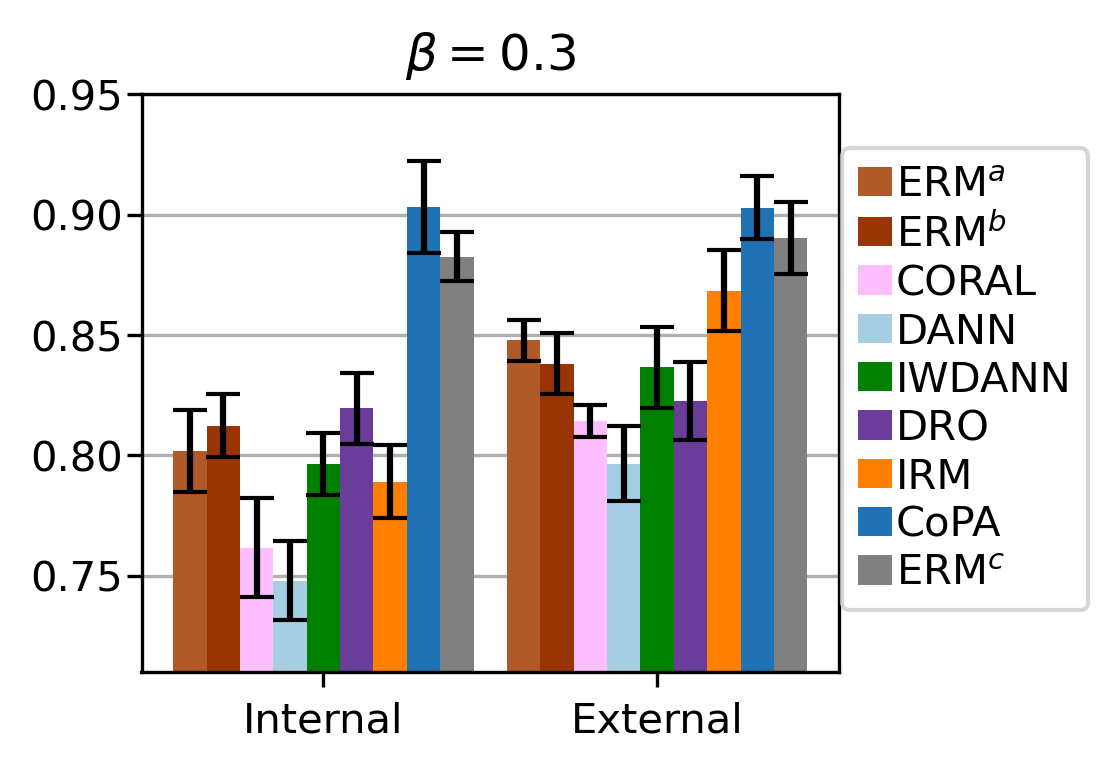}
\end{subfigure}
\caption{F1-score at test site, \emph{multiple} training sites.~\emph{$Y{\leftarrow}Z$}
}\label{}
\end{figure}

\begin{figure}[ht]
\centering
\begin{subfigure}[t]{\linewidth}
\caption{2-dim,\;\; $\text{ERM}^a$: input=X,\;\; $\text{ERM}^b$: input=X,Z}
\centering
\includegraphics[width=.7\linewidth]{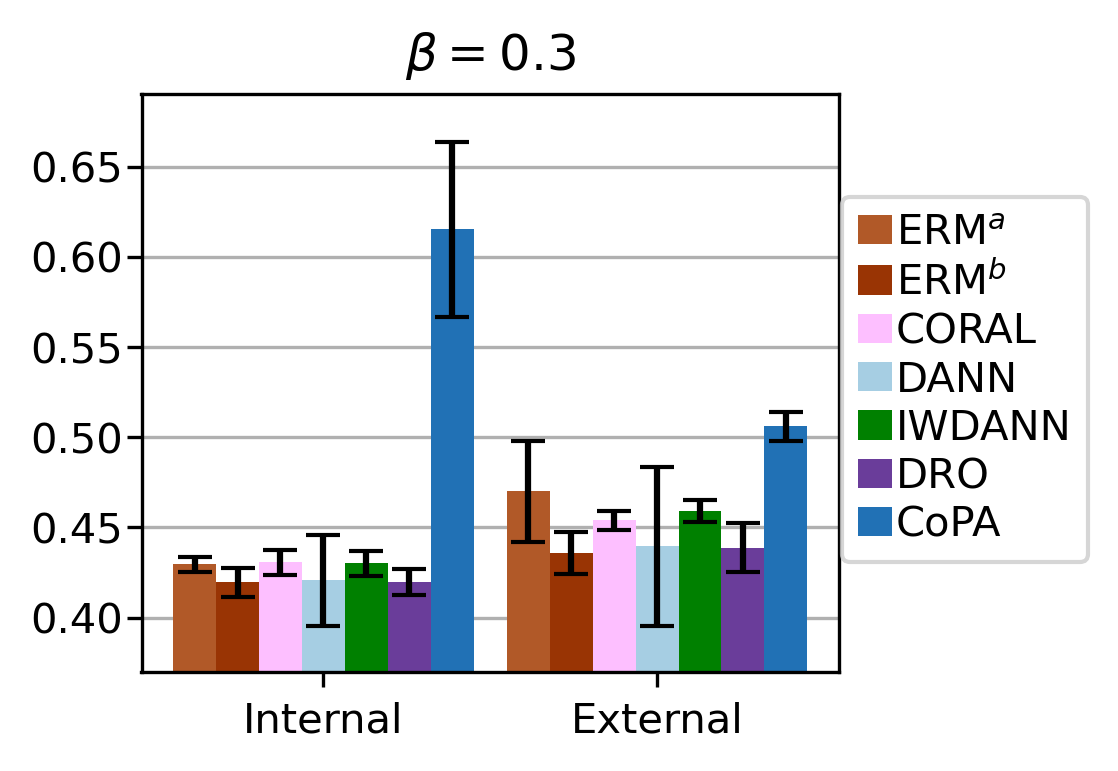}
\end{subfigure}
\begin{subfigure}[t]{\linewidth}
\caption{CMNIST,\;\; $\text{ERM}^c$: greyscale input}
\centering
\includegraphics[width=.7\linewidth]{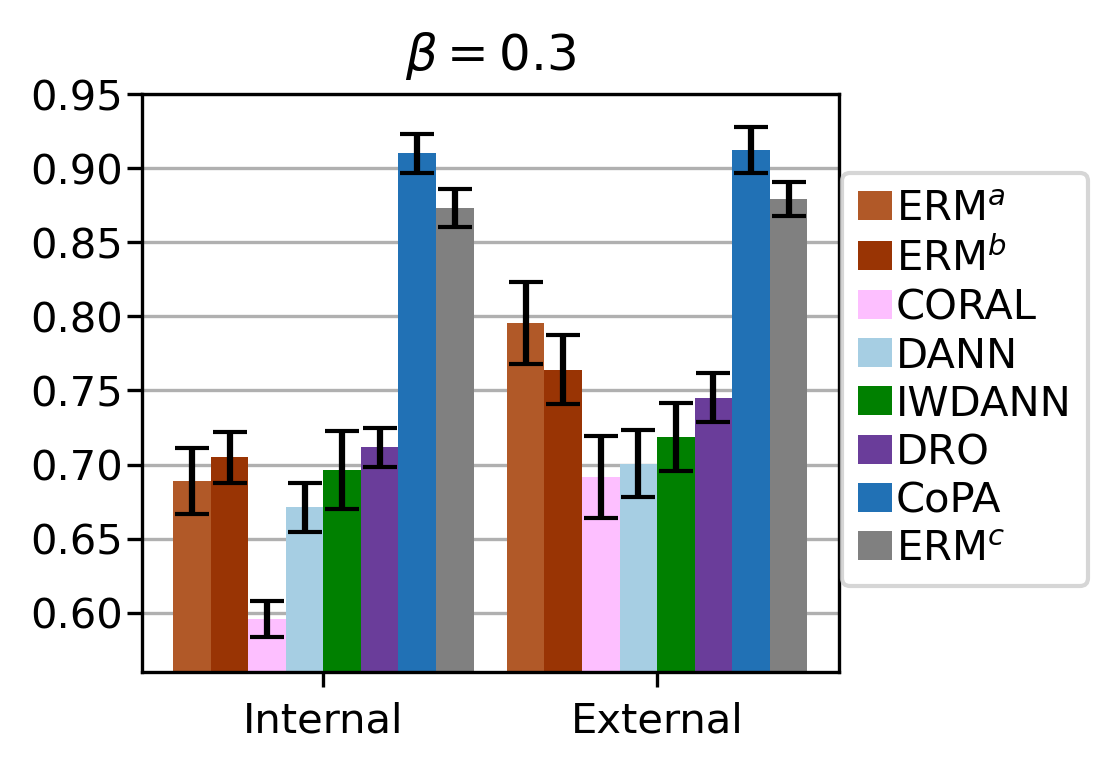}
\end{subfigure}
\caption{F1-score at test site, a \emph{single} training site.~\emph{$Y{\leftarrow}Z$}
}\label{}
\end{figure}

\begin{figure}[ht]
\centering
\begin{subfigure}[t]{\linewidth}
\caption{2-dim,\;\; $\text{ERM}^a$: input=X,\;\; $\text{ERM}^b$: input=X,Z}
\centering
\includegraphics[width=.7\linewidth]{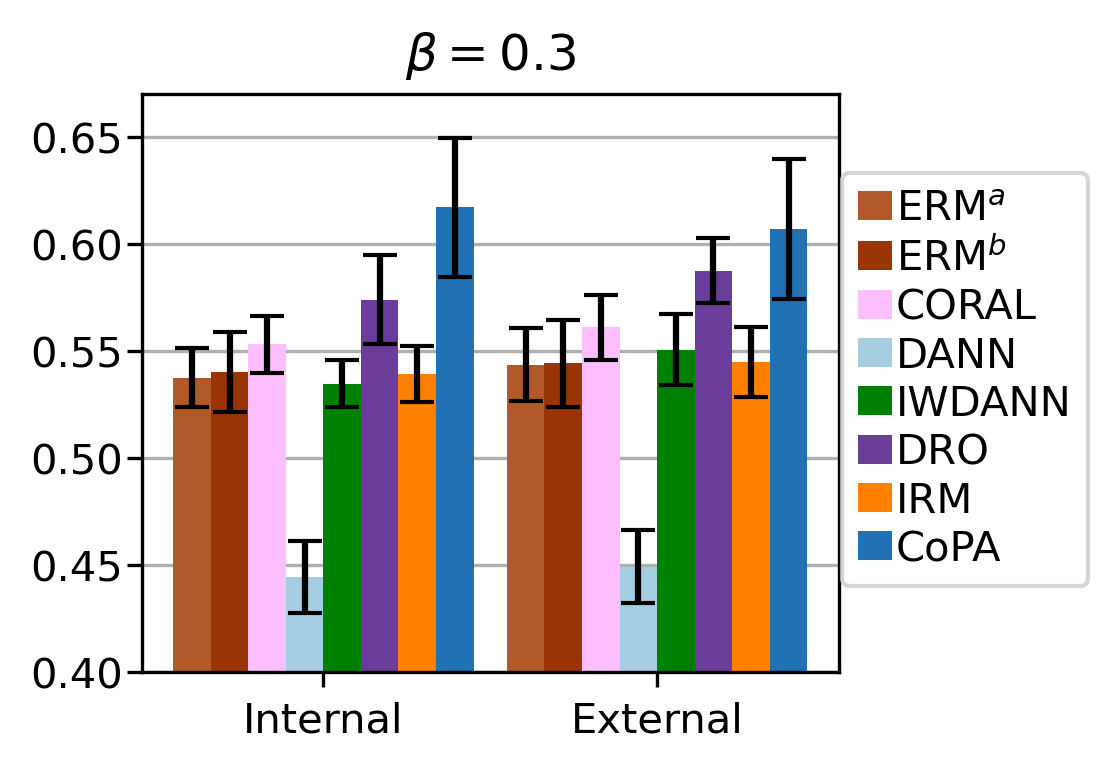}
\end{subfigure}
\begin{subfigure}[t]{\linewidth}
\caption{CMNIST,\;\; $\text{ERM}^c$: greyscale input}
\centering
\includegraphics[width=.7\linewidth]{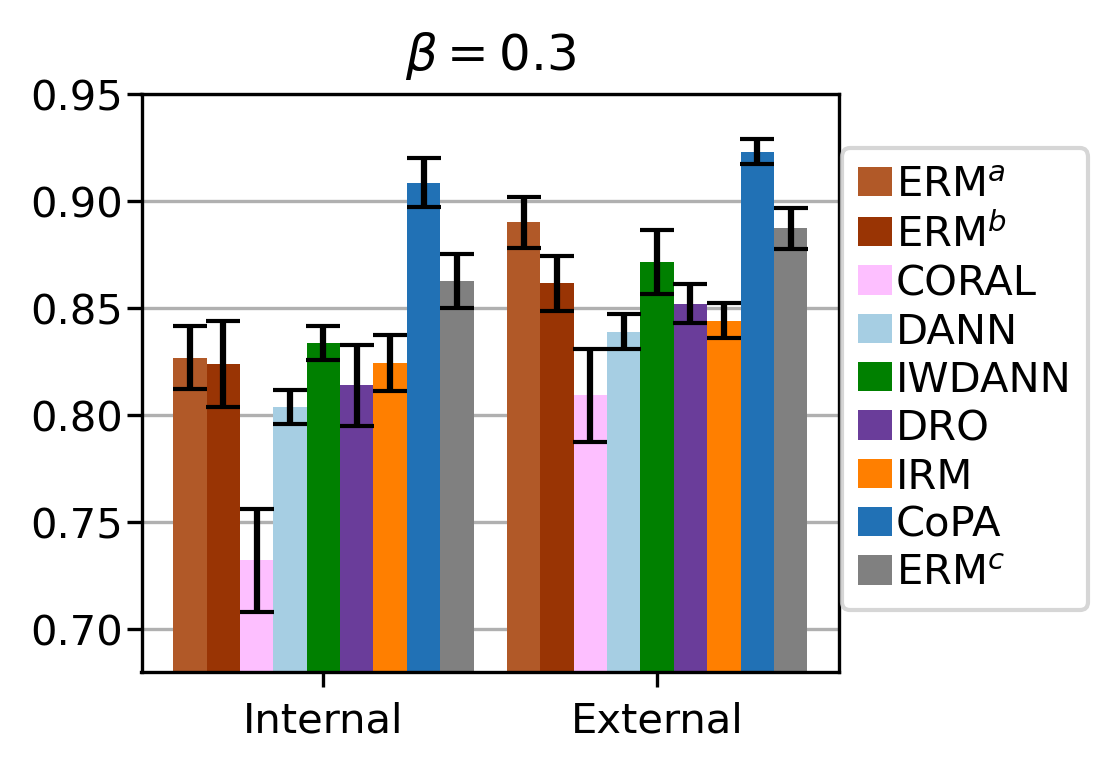}
\end{subfigure}
\caption{F1-score at test site, \emph{multiple} training sites.~\emph{$Y{\rightarrow}Z$}
}\label{}
\end{figure}

\begin{figure}[ht]
\centering
\begin{subfigure}[t]{\linewidth}
\caption{2-dim,\;\; $\text{ERM}^a$: input=X,\;\; $\text{ERM}^b$: input=X,Z}
\centering
\includegraphics[width=.7\linewidth]{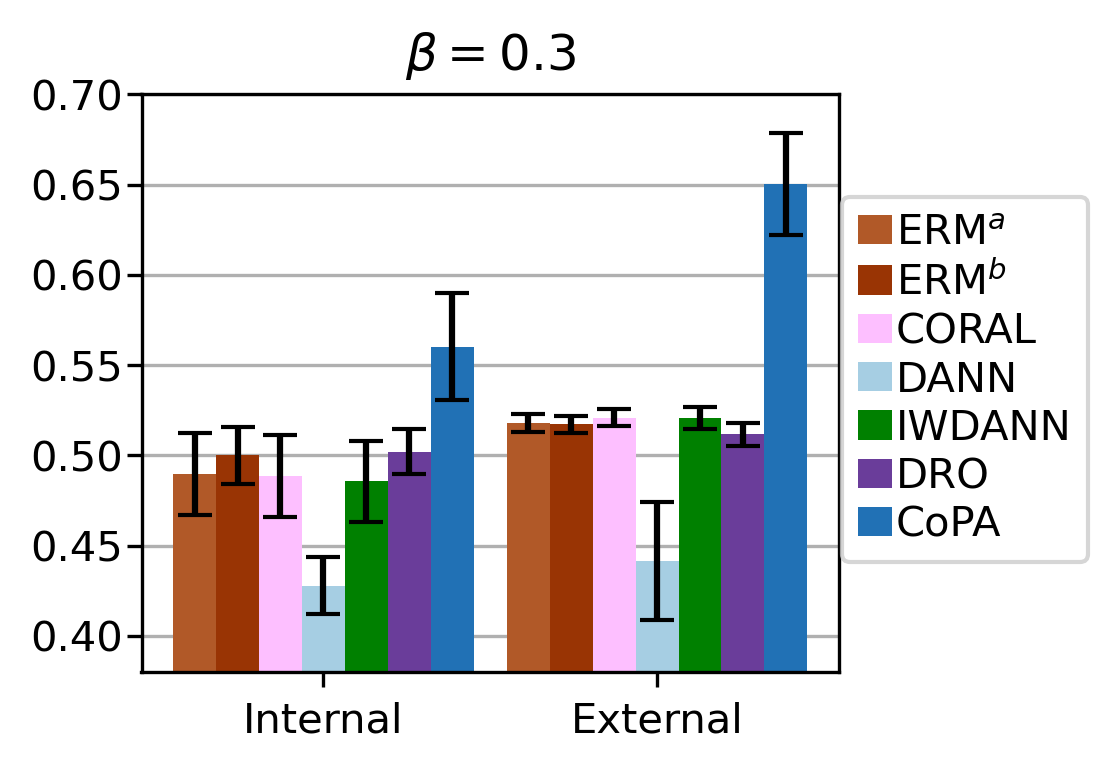}
\end{subfigure}
\begin{subfigure}[t]{\linewidth}
\caption{CMNIST,\;\; $\text{ERM}^c$: greyscale input}
\centering
\includegraphics[width=.7\linewidth]{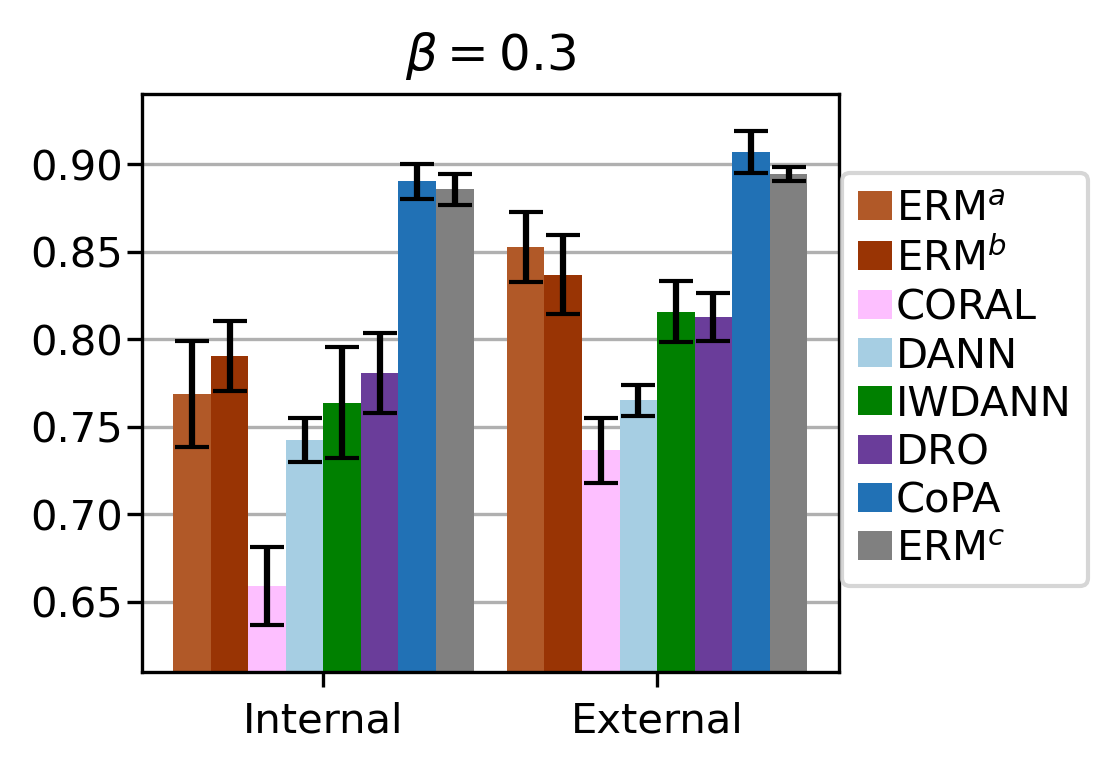}
\end{subfigure}
\caption{F1-score at test site, a \emph{single} training site.~\emph{$Y{\rightarrow}Z$}
}\label{}
\end{figure}

\begin{figure}[ht]
\centering
\includegraphics[width=\linewidth]{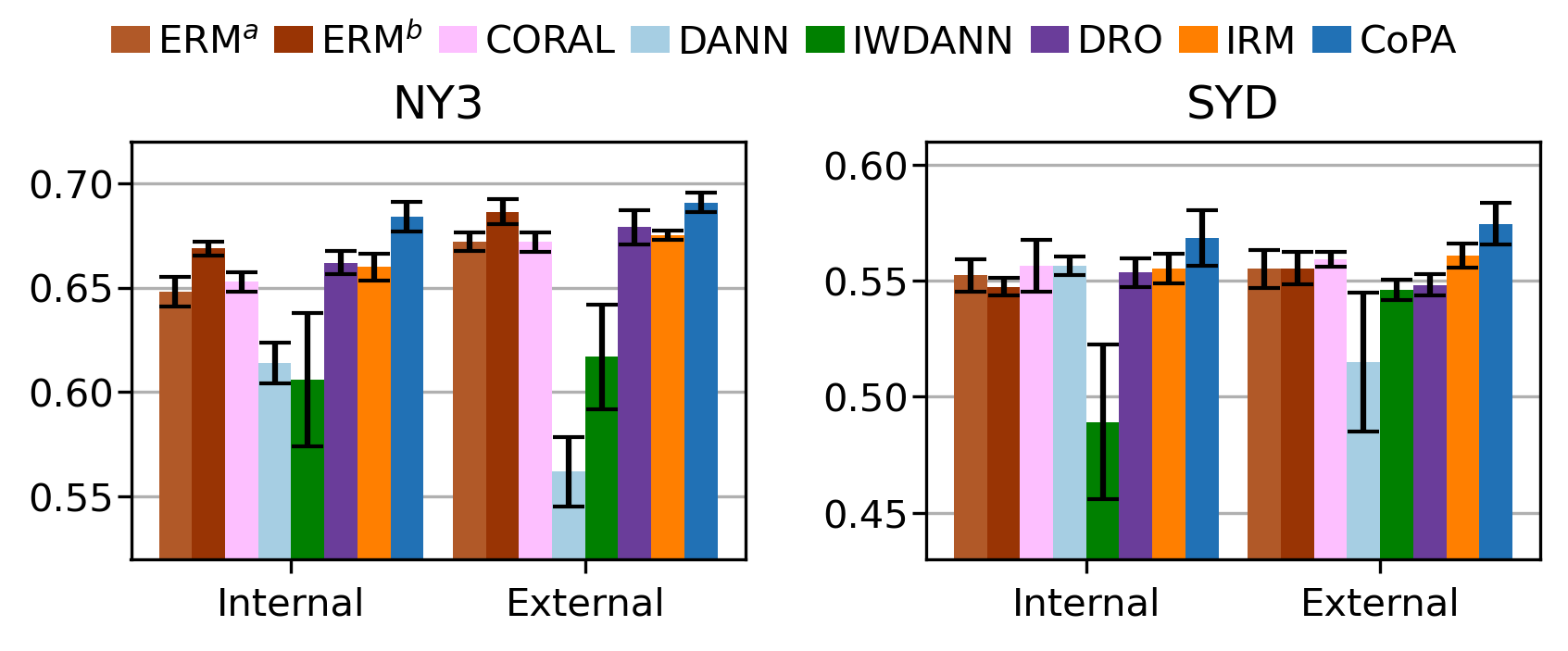}
\caption{F1-score at ISIC test sites, \emph{multiple} training sites.}
\end{figure}

\begin{figure}[ht]
\centering
\includegraphics[width=.7\linewidth]{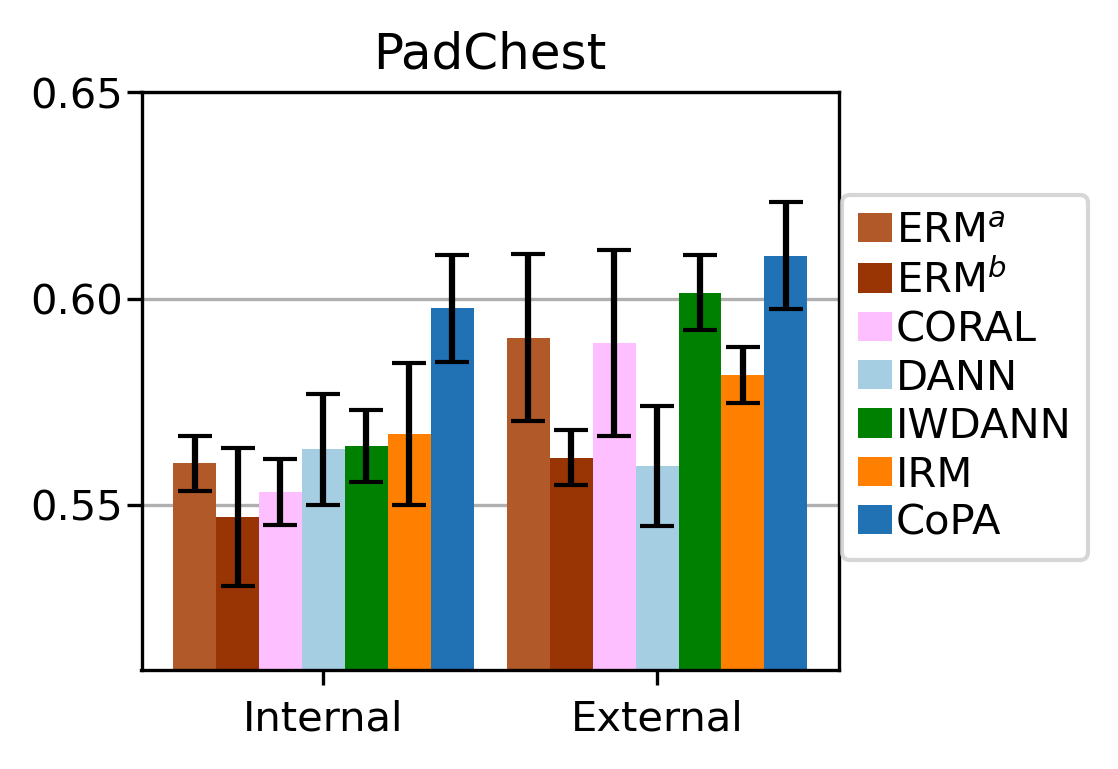}
\caption{F1-score at CXR test site, a \emph{single} training site}
\end{figure}

\clearpage
\section{Complete Ablations}
\begin{figure}[ht]
\centering
\begin{subfigure}[t]{.9\linewidth}
\caption{2-dim}
\centering
\includegraphics[width=.7\linewidth]{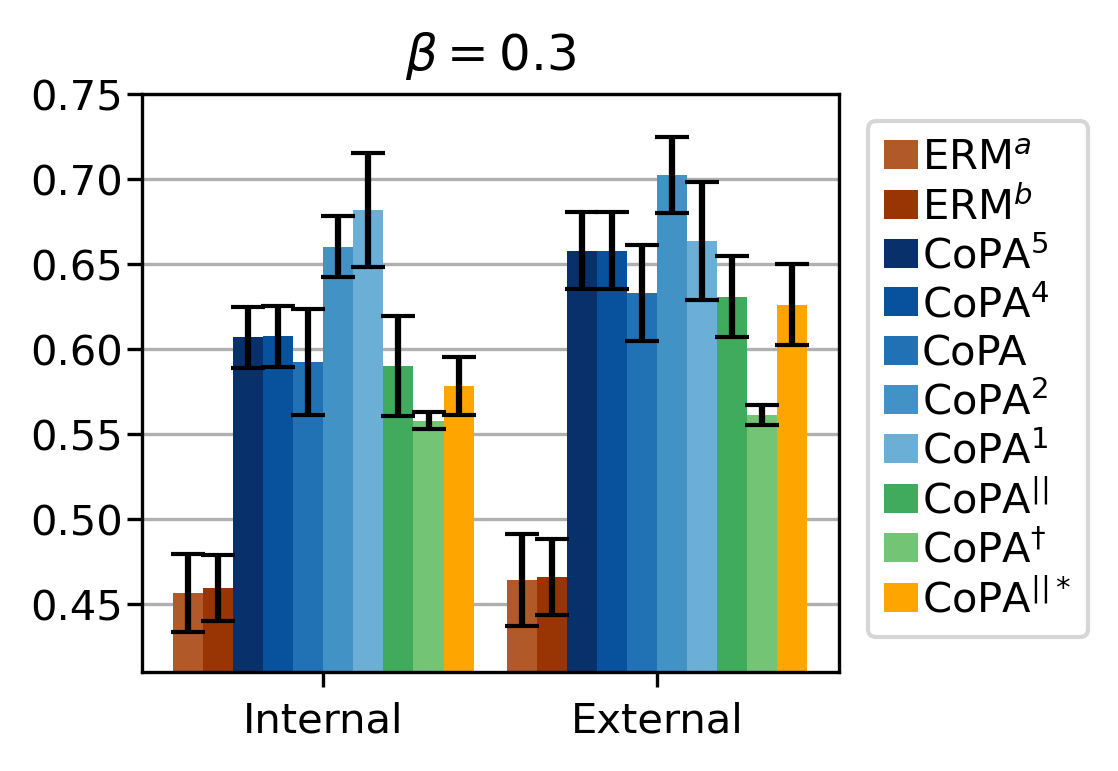}
\end{subfigure}
\begin{subfigure}[t]{\linewidth}
\caption{CMNIST}
\centering
\includegraphics[width=.7\linewidth]{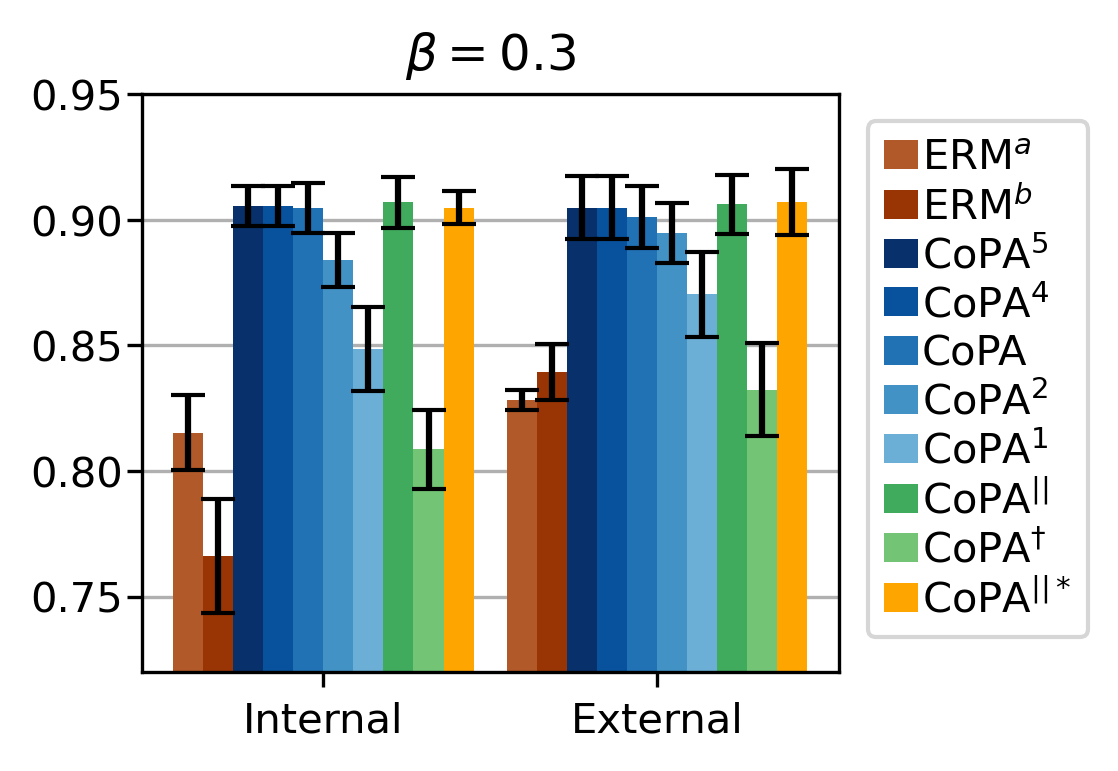}
\end{subfigure}
\caption{Ablation on synthetic data. Test F1-score. $Y{\leftarrow}S{\rightarrow}Z$
}\label{}
\end{figure}

\begin{figure}[ht]
\centering
\begin{subfigure}[t]{.9\linewidth}
\caption{2-dim}
\centering
\includegraphics[width=.7\linewidth]{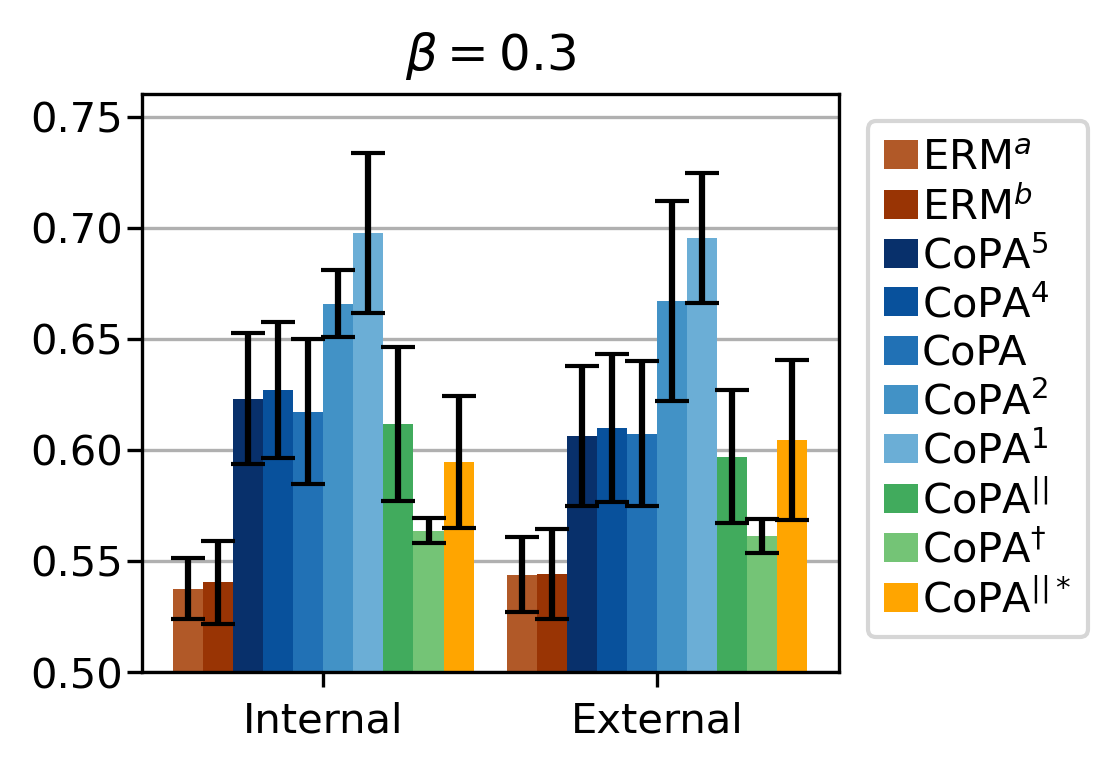}
\end{subfigure}
\begin{subfigure}[t]{.9\linewidth}
\caption{CMNIST}
\centering
\includegraphics[width=.7\linewidth]{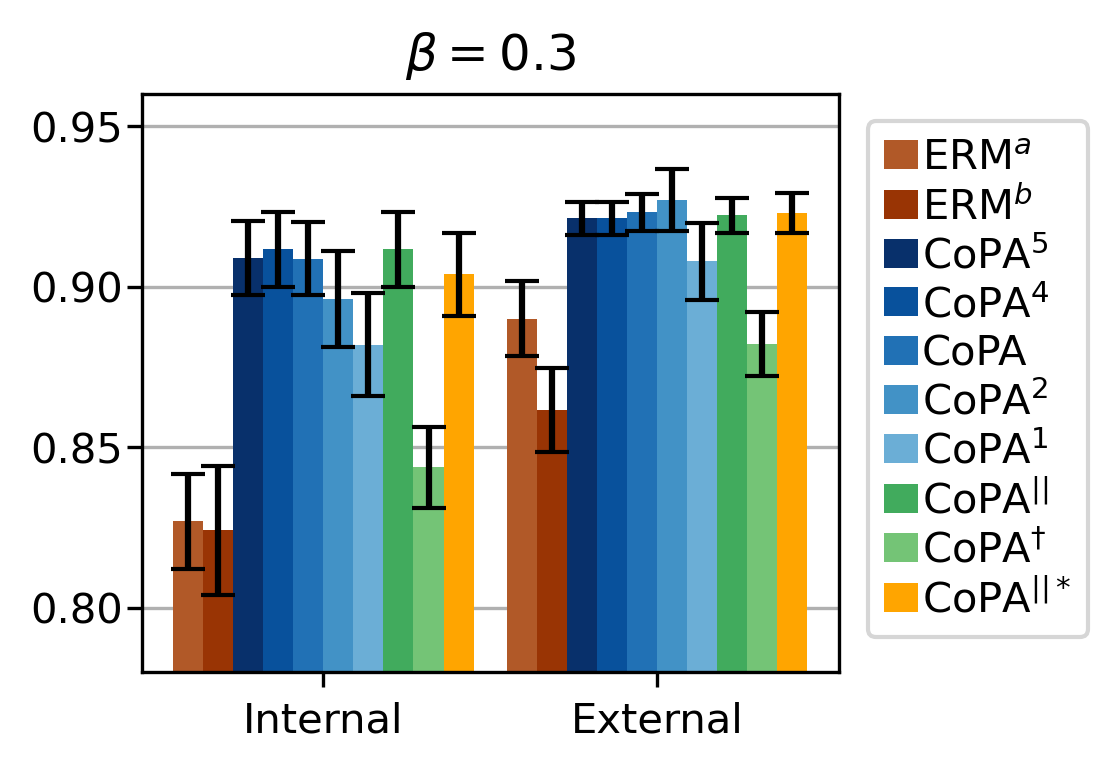}
\end{subfigure}
\caption{Ablation on synthetic data. Test F1-score. $Y{\rightarrow}Z$
}\label{}
\end{figure}

\begin{figure}[ht]
\centering
\begin{subfigure}[t]{.9\linewidth}
\caption{2-dim}
\centering
\includegraphics[width=.7\linewidth]{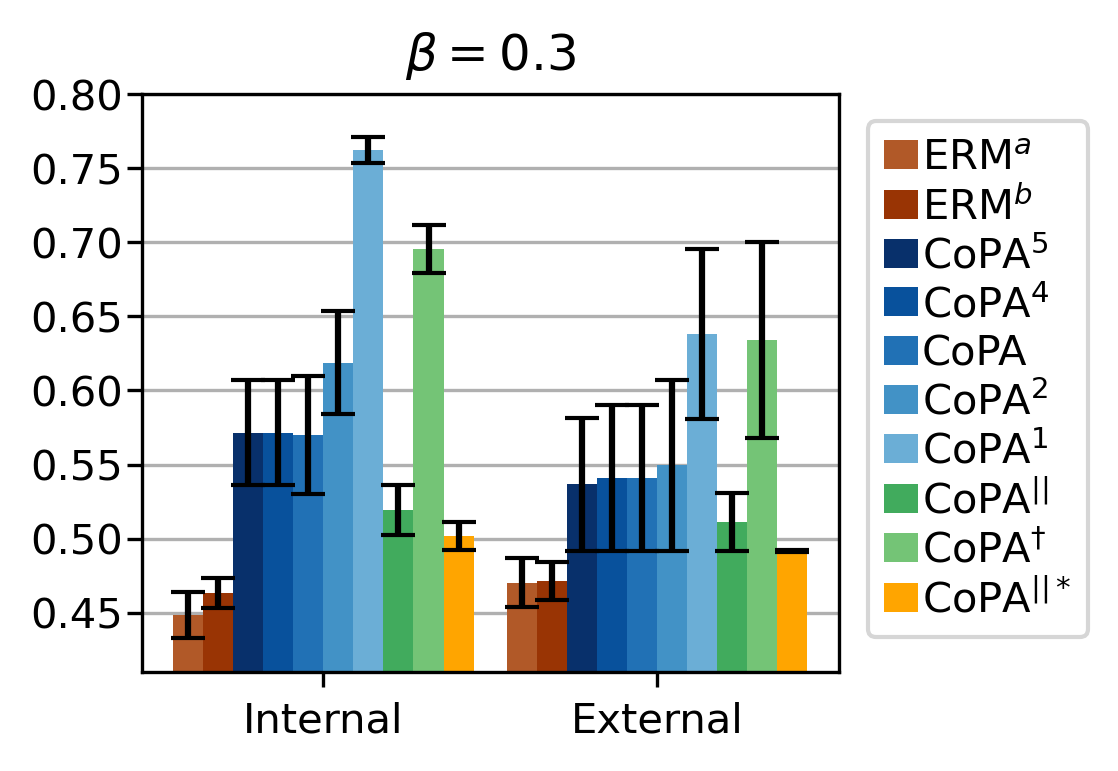}
\end{subfigure}
\begin{subfigure}[t]{.9\linewidth}
\caption{CMNIST}
\centering
\includegraphics[width=.7\linewidth]{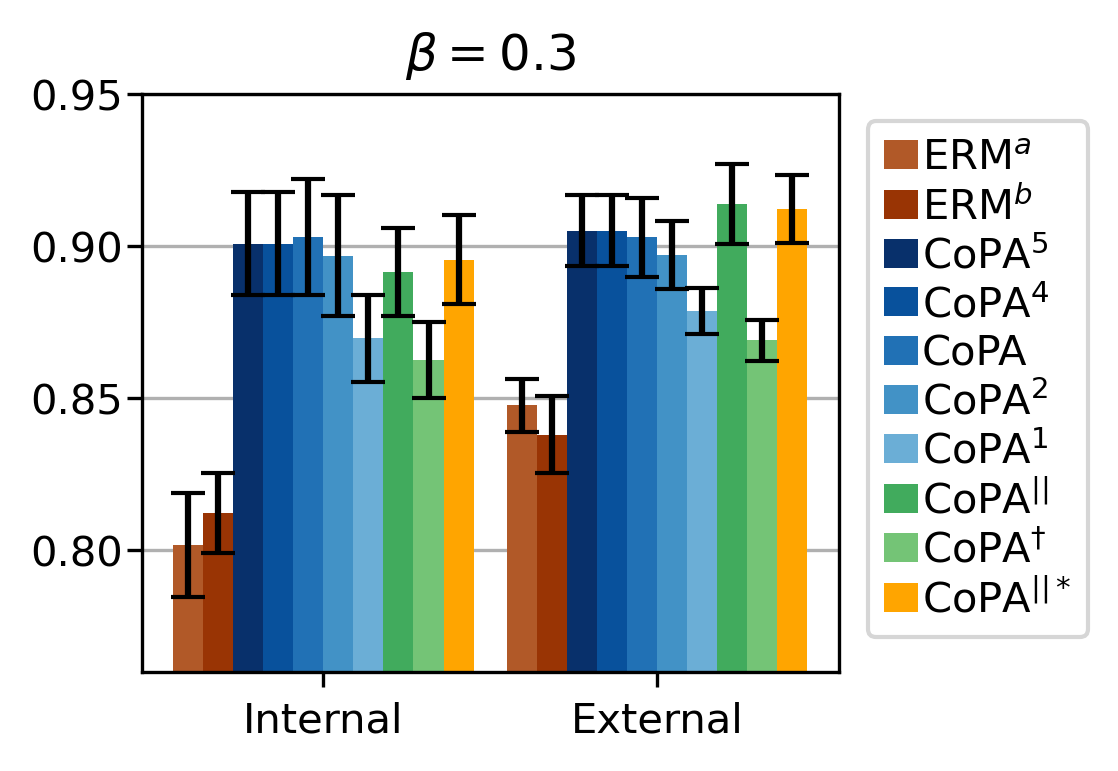}
\end{subfigure}
\caption{Ablation on synthetic data. Test F1-score. $Y{\leftarrow}Z$
}\label{}
\end{figure}

\begin{figure}[ht]
\centering
\includegraphics[width=\linewidth]{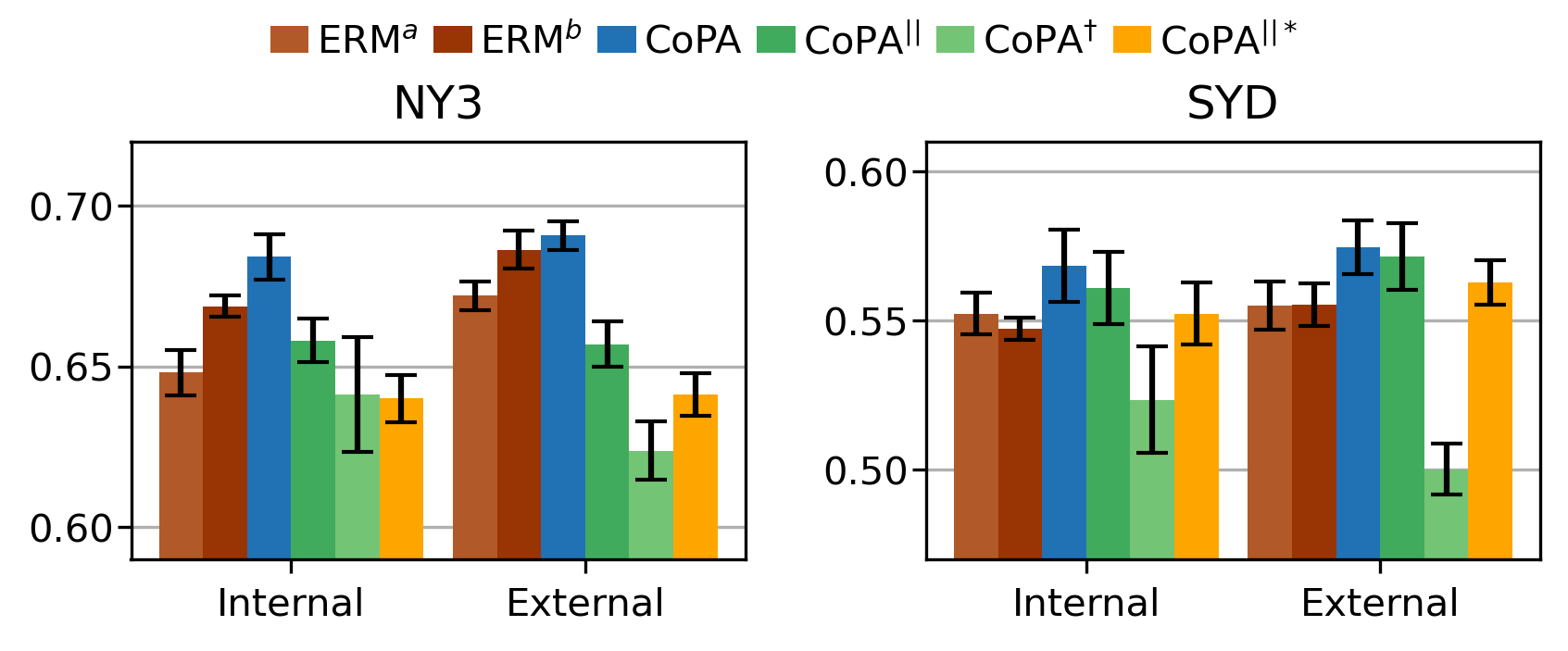}
\caption{Ablation on ISIC data. Test F1-score.
}\label{}
\end{figure}

\begin{figure}[ht]
\centering
\includegraphics[width=.7\linewidth]{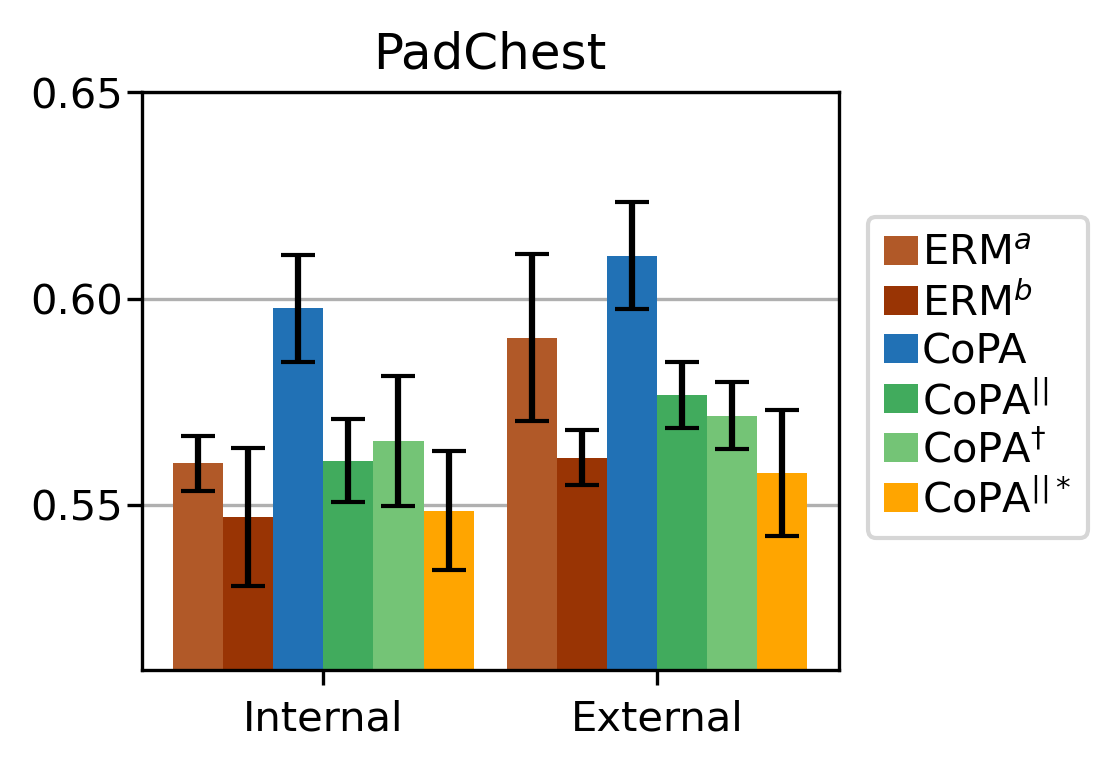}
\caption{Ablation on CXR data. Test F1-score.
}\label{}
\end{figure}

\section{Examples of CMNIST, ISIC, and CXR data}\label{app:data_example}

\subsection{CMNIST}
\begin{figure}[ht]
\centering
\includegraphics[width=\linewidth]{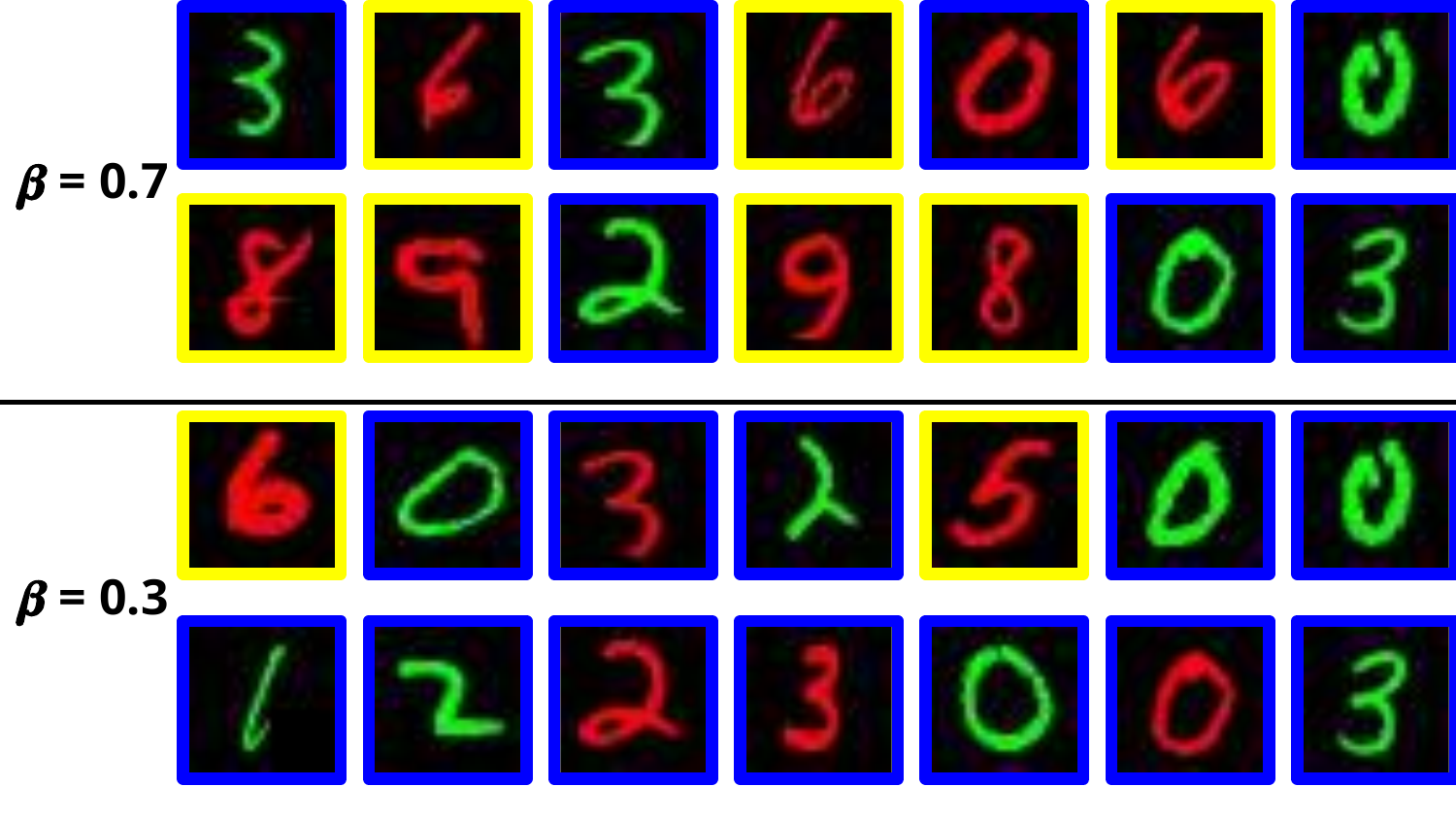}
\caption{CMNIST data from two sites: $\beta{=}0.7$ and $\beta{=}0.3$. Yellow border: $Y{=}1$, blue border: $Y{=}0$.
}\label{fig:viz_cmnist}
\end{figure}

Figure~\ref{fig:viz_cmnist} shows some examples of the CMNIST data from two sites: $\beta{=}0.7$ and $\beta{=}0.3$.
The correlation between red color and $Y{=}1$ is strong when $\beta{=}0.7$ but is very weak when $\beta{=}0.3$.
Besides, there are far fewer images with $Y{=}1$ label in $\beta{=}0.3$ site than in $\beta{=}0.7$ site, indicating a change in the $P(Y|E)$ distribution.

\subsection{ISIC Data}
The skin cancer dataset is from the International Skin Imaging Collaboration (ISIC) archive.
Data from the archive are collected by Memorial Sloan Kettering Cancer Center, Medical University of Vienna, Hospital Clinic de Barcelona, Melanoma Institute Australia, the University of Queensland, and Boston University at different points in time.
There are about 70k data samples in total.
Each data sample consists of an input image $X$, a binary target label $Y$ (having melanoma or not) and three confounding variables $Z$ (\textit{Age}, \textit{Anatomical Site}, \textit{Sex}).
Table~\ref{tab:isic_z_list} shows the list of values of \textit{Anatomical Site}.
\begin{table}[ht]
\centering
\begin{tabular}{lr}
    \textit{Anatomical Site} & Frequency (\%) \\
    \midrule
    anterior torso & 17.94 \\
    head/neck & 9.83 \\
    lateral torso & 1.24 \\
    lower extremity & 19.91 \\
    oral/genital & 0.27 \\
    palms/soles & 1.13 \\
    posterior torso & 16.67 \\
    upper extremity & 12.08
\end{tabular}
\caption{\textit{Anatomical Site} as a confounding variable ($Z$)
}\label{tab:isic_z_list}
\end{table}

\begin{figure*}[ht]
\centering
\includegraphics[width=\linewidth]{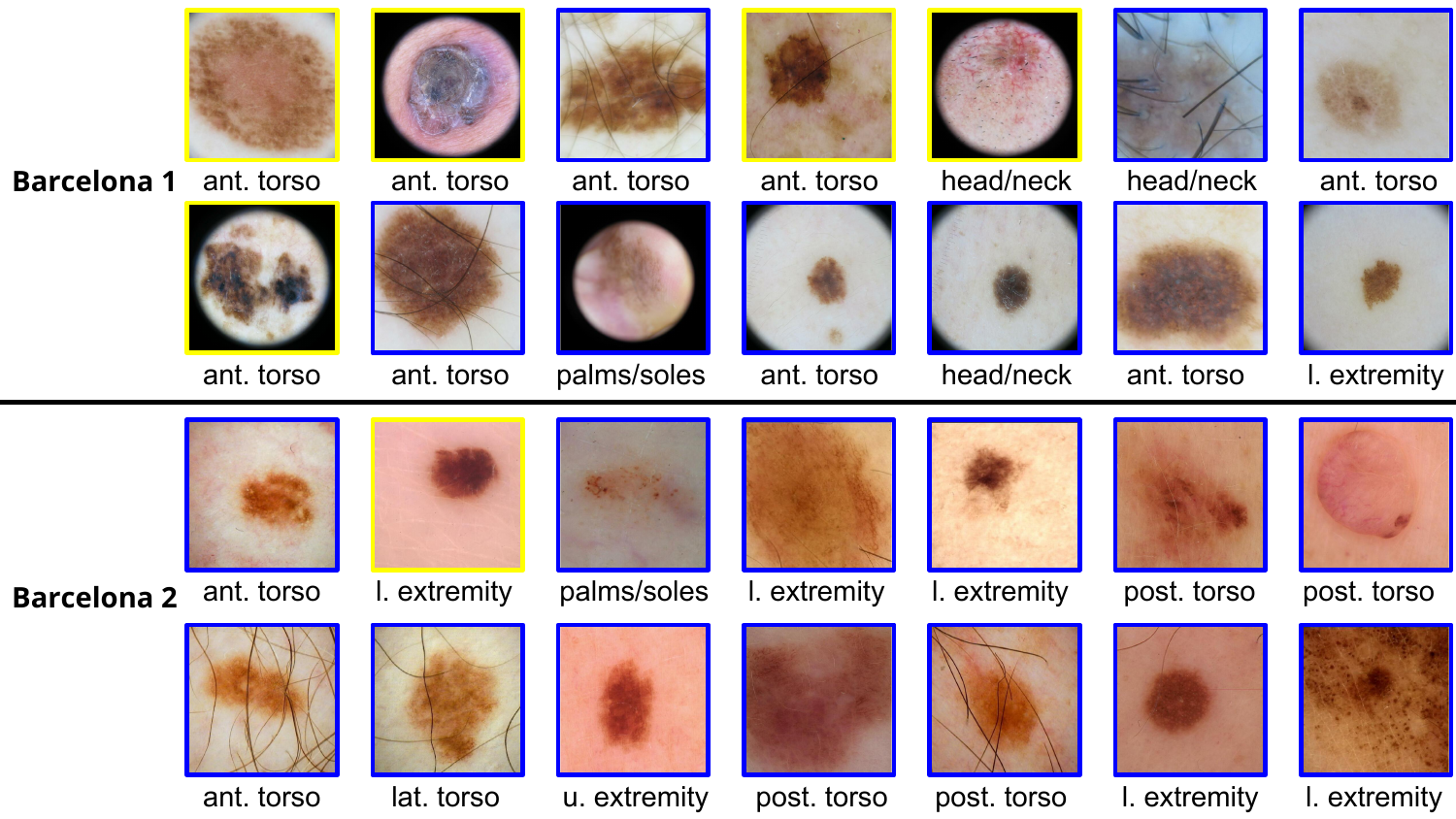}
\caption{ISIC data from two sites: \textit{Barcelona 1} (\textit{BCN1}) and \textit{Barcelona 2} (\textit{BCN2}).
Yellow border: $Y{=}1$, blue border: $Y{=}0$.
The captions under the images indicate the \textit{Anatomical Site} where the images were taken.
}\label{fig:viz_isic}
\end{figure*}

Figure~\ref{fig:viz_isic} shows the ISIC data from two sites: \textit{BCN1} and \textit{BCN2}.
There are also far fewer images with $Y{=}1$ label in \textit{BCN2} than in \textit{BCN1}, indicating a change in the $P(Y|E)$ distribution.

\subsection{Chest X-Ray (CXR) Data}

\begin{figure*}[ht]
\centering
\includegraphics[width=\linewidth]{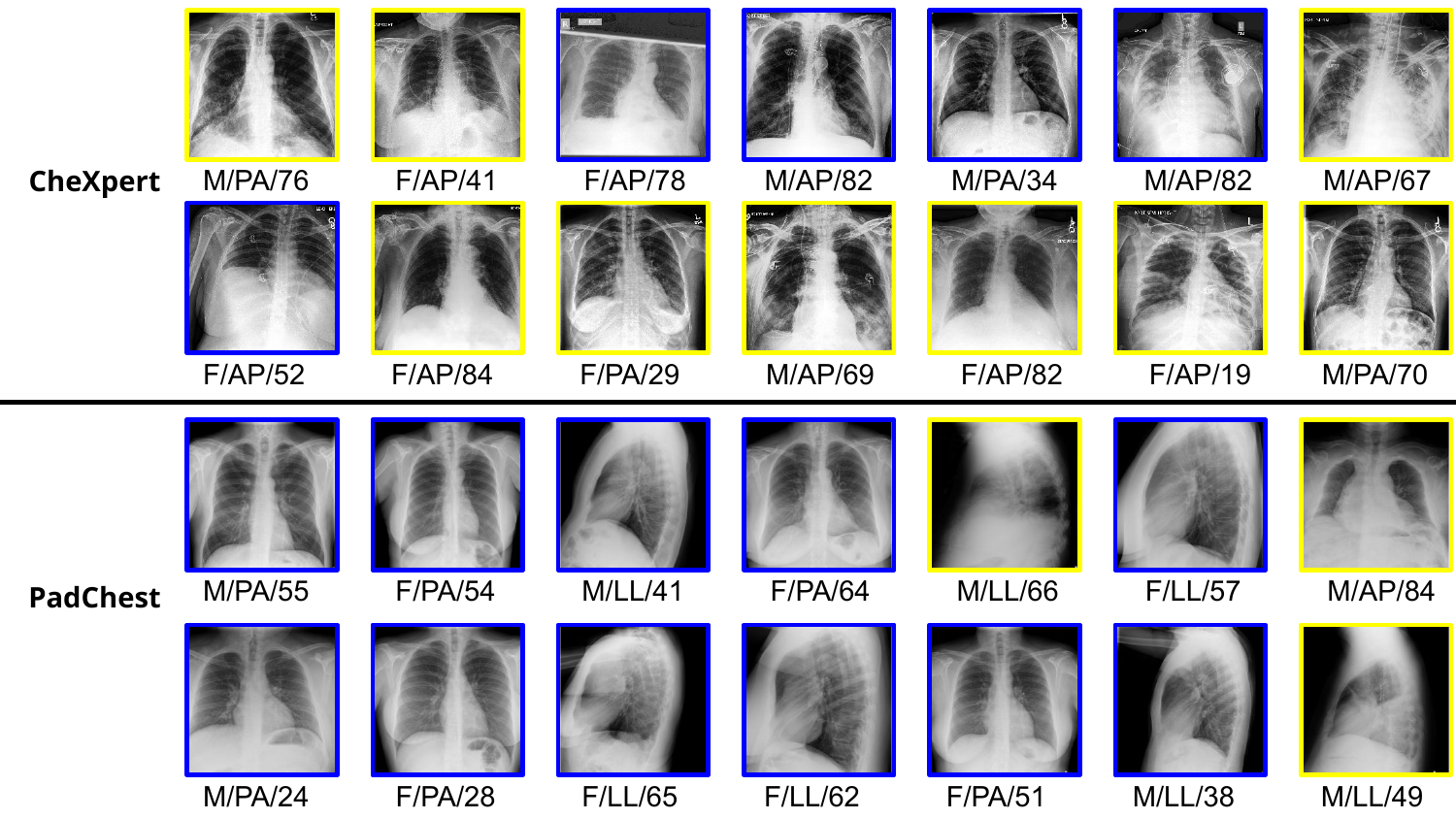}
\caption{Data from two chest X-Ray datasets: CheXpert and PadChest.
Yellow border: $Y{=}1$, blue border: $Y{=}0$.
The captions under the images indicate the \textit{Sex}/\textit{Projection}/\textit{Age} values.
}\label{fig:viz_cxr}
\end{figure*}

Figure~\ref{fig:viz_cxr} shows data from two datasets: CheXpert and PadChest.

\end{document}